\PassOptionsToPackage{svgnames}{xcolor}
\documentclass[]{fairpreprint}

\author{Maximilian Nickel}
\affiliation{FAIR at Meta}
\date{\today}
\correspondence{Maximilian Nickel at \email{maxn@meta.com}}%
\abstract{Rapid model validation via the train-test paradigm has been a key driver for the breathtaking progress in machine learning and AI.\@
However, modern AI systems often depend on a combination of tasks and data collection practices that violate all assumptions ensuring test validity.
Yet, without rigorous model validation we cannot ensure the intended outcomes of deployed AI systems, including positive social impact, nor continue to advance AI research in a scientifically sound way. 
In this paper, I will show that for widely considered inference settings in complex social systems the train-test paradigm does not only lack a justification but is indeed invalid for any risk estimator, including counterfactual and causal estimators, with high probability. 
These formal impossibility results highlight a fundamental epistemic issue, i.e., that for key tasks in modern AI we cannot know whether models are valid under current data collection practices. 
Importantly, this includes variants of both recommender systems and reasoning via large language models, and neither na{\"\i}ve scaling nor limited benchmarks are suited to address this issue.
I am illustrating these results via the widely used \textsc{MovieLens} benchmark and conclude by discussing the implications of these results for AI in social systems, including possible remedies such as participatory data curation and open science.%
}

\usepackage{amsmath}
\usepackage{amsthm}
\usepackage{thmtools}
\usepackage[ragged]{footmisc}

\makeatletter
\renewcommand\@makefntext[1]{\leftskip=0em\hskip0em\@makefnmark#1}
\makeatother

\usepackage{tikz}
\usetikzlibrary{external}
\tikzsetexternalprefix{figs/tmp/}
\usetikzlibrary{positioning,fit,arrows,matrix,decorations.pathmorphing,shapes,patterns}

\tikzexternaldisable

\usepackage{pgfplots}
\pgfplotsset{compat=1.18}

\newcommand{\isneurips}[2]{\IfPackageLoadedTF{neurips_2024}{#1}{#2}}

\isneurips{%
    \title{{No free delivery service}\\ \normalfont Epistemic limits of passive data collection\\ in complex social systems}

    \newcommand{\figtitle}[1]{\textbf{#1}}
    \newcommand{\citestyle}{numeric-comp}
    \newcommand{\proofindent}{0em}
    \usepackage[scaled=0.9]{helvet}
}{
    \title{No Free Delivery Service\\ \large{\normalfont\sffamily Epistemic limits of passive data collection in complex social systems}}

    \newcommand{\figtitle}[1]{{\sffamily\textbf{#1}}}
    \newcommand{\citestyle}{authoryear-comp}
    \newcommand{\proofindent}{1.5em}
}

\usepackage[natbib=true,backend=biber,style=,maxbibnames=99,sortcites=false,maxcitenames=1]{biblatex}
\usepackage{subcaption}
\usepackage{enumitem}

\setlist[description]{font=\normalfont\itshape,leftmargin=3ex}

\newcommand{\R}{\mathbb{R}}
\newcommand{\draw}{\sim}
\newcommand{\dist}[1]{\mathsf{#1}}
\renewcommand{\Pr}{\mathbb{P}}
\DeclareMathOperator{\E}{\mathbb{E}}
\DeclareMathOperator*{\argmin}{\text{arg\,min}}
\DeclareMathOperator*{\arginf}{\text{arg\,inf}}
\newcommand{\pworlds}{\Set{F}}

\newcommand{\infset}{\ensuremath{(\A, \dset, \tdist, \fdist)}}
\newcommand{\A}{\mathbb{A}}
\newcommand{\hyps}{\Set{H}}
\newcommand{\sdist}{\dist{S}}
\newcommand{\tdist}{\dist{T}}
\newcommand{\fdist}{\dist{F}}
\newcommand{\udist}{\dist{U}}
\newcommand{\trisk}{L^{\tdist}_{fh}}
\newcommand{\PrF}{\Pr_{f~\draw~\fdist}}

\newcommand{\squal}{\theta}
\newcommand{\obsvgraph}{\ensuremath{\Set{S}}}
\newcommand{\mat}[1]{\mathbf{#1}}
\newcommand{\ff}{f}
\newcommand{\mY}{\mat{Y}}
\newcommand{\mF}{\mat{F}}
\newcommand{\mU}{\mat{U}}
\newcommand{\mV}{\mat{V}}
\newcommand{\mQ}{\mat{Q}}
\newcommand{\mR}{\mat{R}}
\newcommand{\mX}{\mat{X}}

\newcommand{\Set}[1]{\mathcal{#1}}
\newcommand{\sset}{\Set{S}}
\newcommand{\dset}{\Set{D}}
\newcommand{\tset}{\Set{T}}

\newcommand{\rank}{\text{rank}}
\newcommand{\kcore}{k\text{-core}}

\newcommand{\proj}{\mathcal{P}}

\newtheoremstyle{mythm}{}{}{\itshape}{}{\itshape}{.}{ }{\thmname{#1}\thmnumber{ #2}\thmnote{ (#3)}}
\newtheoremstyle{mydef}{}{}{\normalfont}{}{\itshape}{.}{ }{\thmname{#1}\thmnumber{ #2}\thmnote{ (#3)}}

\newtheorem{theorem}{Theorem}
\newtheorem*{theorem*}{Theorem}

\newtheorem{corollary}{Corollary}
\newtheorem*{corollary*}{Corollary}
\newtheorem{proposition}{Proposition}
\newtheorem*{proposition*}{Proposition}
\theoremstyle{definition}
\newtheorem{definition}{Definition}

\newtheorem{example}{Example}[]
\newtheorem*{evidence}{Supporting evidence}
\newtheorem{rques}{Research Question}

\declaretheoremstyle[%
  spaceabove=0pt,%
  spacebelow=6pt,%
  headfont=\normalfont\itshape,%
  headindent=\proofindent,%
  qed=\qedsymbol%
]{mystyle} 
\declaretheorem[name={Proof sketch},style=mystyle,unnumbered,
]{prf}

\isneurips{\usepackage{newtxmath}}{}

\Crefname{appendix}{Supplement}{Supplements}
\crefname{appendix}{supp.}{supps.}
\crefname{rques}{RQ}{RQs}

\AtBeginBibliography{\small}

\begin{document}

\maketitle

\isneurips{%
    \begin{abstract}%
    \end{abstract}%
}{}

\begin{figure}[h!]
    \centering
    \hspace*{3em}\resizebox{.85\linewidth}{!}{%
    \input{figs/fig1.tikz}%
}
    \caption{%
        \figtitle{Test validity in complex systems}.
        Given assumptions \(\A\), target distribution \(\tdist\), data set \(\dset \draw \sdist^m\) from a sampling distribution \(\sdist\), and quality metric \(\smash{\squal}\), an inference setting is \emph{test-valid} if the difference between \(\theta\) and the true risk \(\trisk\) can be bounded over the distribution of all possible worlds \(f \draw \fdist\) consistent with \((\A, \dset)\).%
    }
    \vspace{-1ex}
\end{figure}

\section{Introduction}\label{sec:intro}
Model validation, long taken to be ``solved'' via the train-test paradigm, has become one of the central challenges in modern machine learning and artificial intelligence.
In unison with the dramatic increase of their capabilities, AI systems are now supposed to solve tasks of vastly expanded scope, including potentially AI-complete tasks such as open domain question answering, autonomous decision making, and ultimately, artificial general intelligence. Even before the recent triumphs of large language models and deep learning, \citet{Anderson2008-fk} proclaimed ``the end of theory'' and the scientific method being obsolete due to the wonders of big data, large-scale computing, and data mining.
At the same time, it is entirely unclear how to rigorously evaluate the quality of models for these ambitious tasks.
This lack of proper evaluation can then materialize in persistent issues of deployed systems related to generalization, e.g., hallucination~\citep{Ji2022-jy}, out-of-distribution generalization~\citep{Liu2021-pa}, fairness~\citep{Barocas2023-kz}, and generalization to the long-tail~\citep{Feldman2020-br,Feldman2020-rj}. 
Importantly, these issues do not only affect the accuracy of models in a vacuum, but can also affect their social impact if they are deployed in consequential social contexts~\citep{Cheng2021-ri}.
In this paper, I aim to connect the former developments with the latter issues through the lens of epistemology. More concretely, I ask: 
\vspace{-1.85em}
\isneurips{%
    \begingroup
    \advance\leftmargini -2.5em
}{}
\begin{quote} 
\begin{rques}\label{rq1}
Given the ambitious tasks that we ask AI systems to solve and given how we currently collect data, \emph{can we know} whether a model performs well for these tasks?
\end{rques}
\end{quote}
\isneurips{%
    \endgroup
    \vspace{-0.5ex}
}{}
Answering \cref{rq1} positively is central not only for the deployment of machine learning systems, but also for scientific progress within artificial intelligence itself.
After all, knowledge of a model's quality is a prerequisite to detect generalization issues and develop improved models. In deployed systems, a model's predictions are useless --- as good as they might be --- without knowing that they are, in fact, reliable. In social systems, where the consequences of model errors can be severe, having this knowledge is of even greater importance.
Hence, the epistemic question of this work gets to the heart of various debates surrounding AI and its capabilities: How can we understand and measure the true capabilities of modern AI systems, which are so very impressive and yet lacking in fundamental ways at the same time~\citep{Bottou2023-ta}? What can we know about the quality of our models? Are our benchmarks suited to give insights into the intended tasks or do they project a false image of quality? How can we develop systems such that they work for everyone? Will na\"{\i}ve scaling solve all these problems or do we need to invest into entirely new approaches for evaluation within the scope of modern AI?

A prerequisite to answering \cref{rq1} positively is the validity of model validation: Without model validation we can not know wether a model is good or bad and without a valid model validation procedure we can not attain this knowledge.
The almost exclusively used method for model validation in machine learning and AI is the ubiquitous train-test paradigm, i.e., the practice of estimating the generalization performance of a model on a test set distinct from the training set.
Arguably, much of the breathtaking progress in machine learning has been driven by the success of this single experimental paradigm as it allows for the rapid validation and, therefore, improvement of models~\citep{Bottou2015-hs}.
However, it is crucial to note that the train-test paradigm is inherently an inductive method that aims to \emph{infer, not measure}, the generalization error of a model from its error on a test set. 
It is well known --- dating back at least to \citet{Hume1739-hj,Hume1748-in} and formalized in the context of machine learning by \citet{Wolpert1996-hh} --- that it is not possible to justify the validity of such inductive inferences without further assumptions.
This raises the question: is the train-test paradigm still valid for the combination of tasks and data sets considered in modern AI and under what assumptions is this the case?

Importantly, such assumptions should be minimal in terms of ontological commitments, i.e., meet \emph{ontological parsimony} (or minimality), since
(a) model validation results can not provide insights about validity in the real world if they are contingent on strong ontological assumptions
(b) any assumptions that are required to ensure the validity of model validation can not be validated through the same method without circular reasoning. 
In traditional machine learning \emph{settings}, these ontological commitments are placed entirely on the data collection process and, as such, the train-test paradigm is indeed suitable to \emph{validate any model assumption} outside the data collection process.
More concretely, under \emph{active data collection}, i.e., when we actively control the data collection process, we can create large enough test sets that are (approximately) sampled i.i.d.~from the target distribution. Under these conditions, it is well known that the train-test paradigm allows us to validate models simply via their performance on this test set --- \emph{without making any further ontological commitments}. This property is the beauty of the train-test paradigm and what makes it so valuable and successful.

However, domains in modern machine learning have become far too large to be covered via data sets in this active and controlled manner --- the required effort would be prohibitively difficult and costly. 
In lieu, \emph{passive data collection} has become the predominant way to create data sets for modern AI systems. Here, data is collected without intervention from \emph{some social system} that generates data within the domain of interest. 
For instance, rather than meticulously collecting independent samples from all possible facts in a domain, training and validation corpora for QA models are gathered from what has been published on the internet. Similarly, preferences of users are collected over items that a recommender system has pre-selected, rather than sampling them i.i.d.~over all possible user-item pairs. 
Importantly, these sample generating systems need not correspond to the target data generating process, have their own internal dynamics, and are driven by complex interactions of their parts and social processes,
e.g., well-known phenomena such as popularity bias~\citep{Abdollahpouri2019-te}, homophily~\citep{Fabbri2020-om,Liu2023-nk}, or feedback loops~\citep{Chaney2018-vk}. %

Hence, I will ground \cref{rq1} in these conditions of current machine learning practice: \emph{Under passive data collection from a social system, can model validation be valid or not?}
To formalize the social systems with which an AI system interacts, I am taking a \emph{complex systems} perspective and describe them as networks with well-established sampling biases and degree distributions. For these properties, I will show how they affect \emph{necessary conditions} of test validity. %
These results can also be understood as a strengthening of the seminal \emph{No Free Lunch} (NFL) theorems for supervised learning~\citep{Wolpert1996-hh} in the context of social systems. While the NFL theorems show the impossibility of an assumption-free general purpose learning algorithm, a common criticism is that they need to assume an induction-hostile universe, i.e., full ontological neutrality~\citep{Sterkenburg2021-ou}. In practice, where assuming a reasonably induction-friendly universe is common, the NFL theorems have had therefore limited impact. In contrast, the results of this work are grounded in current machine learning practice and considerably stronger: Even for non-trivial assumptions of an induction-friendly universe, model validation can be shown to be invalid when data is collected passively in social systems. 
In other words, there is \emph{no free delivery service} of data for model validation in complex social systems.
To discuss the above results, I will provide a synthesis of results from learning theory, social science, and complex systems --- and combine them with new theoretical and empirical results on the validity of model validation. 
In particular, the \emph{main contributions} of this paper are as follows:
\vspace{-1.85em}
\isneurips{%
\begingroup
\advance\leftmargini -2.5em
}{}
\begin{quote}
\begin{theorem}[Informal]\label{thm:informal1}
    \normalfont
    For passively collected data in complex social systems the train-test paradigm \underline{cannot} be valid under ontological parsimony for the vast majority of the system. This includes widely considered variants of recommender systems and question answering.
\end{theorem}
\setcounter{theorem}{0}
\setcounter{corollary}{1}
\begin{corollary}[Informal]\label{cor:informal1}
    \normalfont
    Na\"{i}ve scaling and limited benchmarks are prohibitively inefficient to address \cref{thm:informal1} and therefore not suited to attain test validity in these scenarios.
\end{corollary}
\setcounter{corollary}{0}
\begin{evidence}
    Theoretical results are supported via experiments on the popular \textsc{MovieLens} benchmark where widely considered recommendation tasks are shown to be test-invalid.
\end{evidence}
\end{quote}
\isneurips{\endgroup}{}

The remainder of this paper proceeds as follows: \Cref{sec:validity,sec:sys} formalize passive data collection in social systems and connect it to test validity. \Cref{sec:results} develops \cref{thm:informal1}, \cref{cor:informal1}, and supporting evidence. \Cref{sec:related,sec:discussion} discuss related work and implications for AI in social systems. %

\section{Passive data collection and inference tasks in social systems}\label{sec:sys}
To construct validation data sets for large-scale domains, there exist currently two main practical approaches: (i) ``scaling'', i.e., indiscriminately collecting as much data as possible from some domain and (ii) manually constructing benchmarks of limited size that probe certain subareas of the domain. In the following, I will focus on formalizing (i) as passive data collection from social systems. \Cref{sec:results} will then show that neither (i) nor (ii) can be solutions to the issues of this paper.    

In sociology, a social system is often considered a pattern of networked interactions that exists between individuals, groups, or institutions \citep{Merriam-Webster-DictionaryUnknown-au}.
For the purposes of this paper, I will consider a \emph{social system} to be a pair \((f, \sdist)\) where \(f : \Set{X} \to \Set{Y}\) is a \emph{possible world of interactions} such that \(\Set{X} = \Set{X}_1 \times \cdots \times \Set{X}_n\) denotes the domain of interactions, \(\Set{Y}\) denotes the set of outcomes (or labels) of an interaction, and \(\sdist : \Set{X} \to [0,1]\) denotes the \emph{sampling distribution} of the system over interactions.
Within this framework, \emph{passive data collection} refers to sampling directly from \(\sdist\). This is in contrast to \emph{active data collection} where we would aim to sample directly from the \emph{target distribution}  \(\tdist : \Set{X} \to [0,1]\) for an inference task, e.g., via simple random sampling, stratified sampling, etc. 

In complex social systems, \(\sdist\) is driven by social processes that lead to two characteristic properties of samples: (i) they are \emph{biased} and (ii) they follow \emph{heavy-tailed} or \emph{power-law} distributions. 
The earliest work on (ii) is due to \citet{Simon1955-nb}, and has independently been discovered in multiple contexts.
In fact, (ii) can often be understood as a consequence of (i), e.g., popularity bias leading to power-law distributions in social networks~\citep{Barabasi1999-ua,Papadopoulos2012-dv}. See also \cref{app:complex} for further discussion of these properties.

In the remainder, I will therefore focus on the presence of heavy-tailed distributions in \(\sset\) to understand how this ubiquitous property of social systems affects test validity.
For this purpose, I will first introduce the concept of a sample graph, i.e., the observed interactions that we receive from \(\sdist\):
\begin{definition}[Sample graph]\label{def:samplegraph}
    A data set \(\obsvgraph \draw \sdist^m \subset \Set{X}_1 \times \Set{X}_2\) of observed interactions induces a bipartite \emph{sample graph} \(G = (\Set{X}_1, \Set{X}_2, \obsvgraph)\) between entities of \(\Set{X}_1\) and \(\Set{X}_2\) where an edge indicates that the corresponding interaction has been observed. In the following, I will use \(\obsvgraph\) and \(G\) interchangeably.
\end{definition} 
For higher arity relations, \cref{def:samplegraph} can easily be generalized to hypergraphs. For simplicity, I will focus on bipartite graphs in the following.
In sample graphs, the heavy-tailed property of complex systems materializes then through their degree distribution. While the exact nature of these distributions is disputed \citep{Broido2019-dj}, I will follow \citet{Voitalov2018-jc} and assume that node degrees in \(\obsvgraph\) follow a \emph{regularly-varying power-law distribution}.
Based on this observation, \emph{passive data in complex social systems} will then refer to the following:
\begin{definition}[Passive data in complex social systems]
Let \(\obsvgraph \draw \sdist^m\) be a sample graph drawn from sampling distribution \(\sdist\). Let \(K_1, K_2\) denote random variables that model the degree distribution in \obsvgraph~of nodes in \(\Set{X}_1\) and \(\Set{X_2}\), respectively. For passively collected data from complex social systems, I will then assume that \(K_1,K_2\) follow regularly-varying power-law distributions, i.e., 
\[\Pr(K_1 > k) = u_1(k)k^{-\alpha_1} \quad \text{and}\quad \Pr(K_2 > k) = u_2(k)k^{-\alpha_2}\]
where \(\alpha_i > 0\) are the tail indices and \(u_i\) are slowly varying functions such that \({\lim_{x \to \infty} u(rx) / u(x) = 1}\) for any \(r > 0\). Higher arity relations are defined analogously.
Next, I will show how passive data in social systems materializes in key inference settings (see also \cref{tab:tasks}).
\end{definition}
\begin{table}[t]
    \caption{\figtitle{Inference settings based on passive data collection in complex social systems.}}\label{tab:tasks}
    \resizebox{\linewidth}{!}{
    \isneurips{%
        \begin{tabular}{p{6em}lp{7em}p{14em}p{11.5em}}
    }{%
        \small
        \begin{tabular}{p{6.5em}lp{7.5em}p{14.5em}p{11.5em}}
    }
        \toprule 
        & \textbf{Domain} \(\Set{X}\) & \textbf{Possible world} \(f\) & \textbf{Sample distribution} \(\sdist\) & \textbf{Target distribution} \(\tdist\) \\
        \midrule
        Recommender systems & 
        \(\Set{U} \times \Set{I}\) & 
        User preferences & 
        Probability of user interacting with item, heavy-tailed in \(\Set{U}\) and \(\Set{I}\) & 
        Uniform,\newline \(p_T(u, i) = 1 / |\Set{U} \times \Set{I}|\) 
        \\
        \cmidrule(r){1-1} \cmidrule(lr){2-2} \cmidrule(lr){3-3} \cmidrule(lr){4-4} \cmidrule(l){5-5}
        Symbolic\newline reasoning & 
        \(\Set{S} \times \Set{P} \times \Set{O}\) &
        Truth value\newline of factoids &
        Probability of observing factoid, heavy-tailed in \(\Set{S}\), \(\Set{P}\), and \(\Set{O}\) &
        Uniform,\newline \(p_T(s, p, o) = 1 / |\Set{S} \times \Set{P} \times \Set{O}|\)
        \\
        \bottomrule
    \end{tabular}}
\end{table}
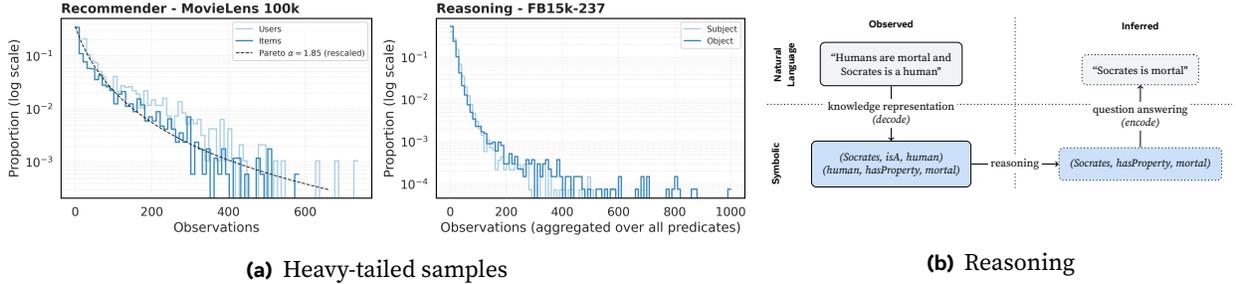
\begin{figure}[t]
    \vspace{-1ex}
    \begin{subfigure}[T]{.6\linewidth}
        \centering
        \includegraphics[width=.5\linewidth]{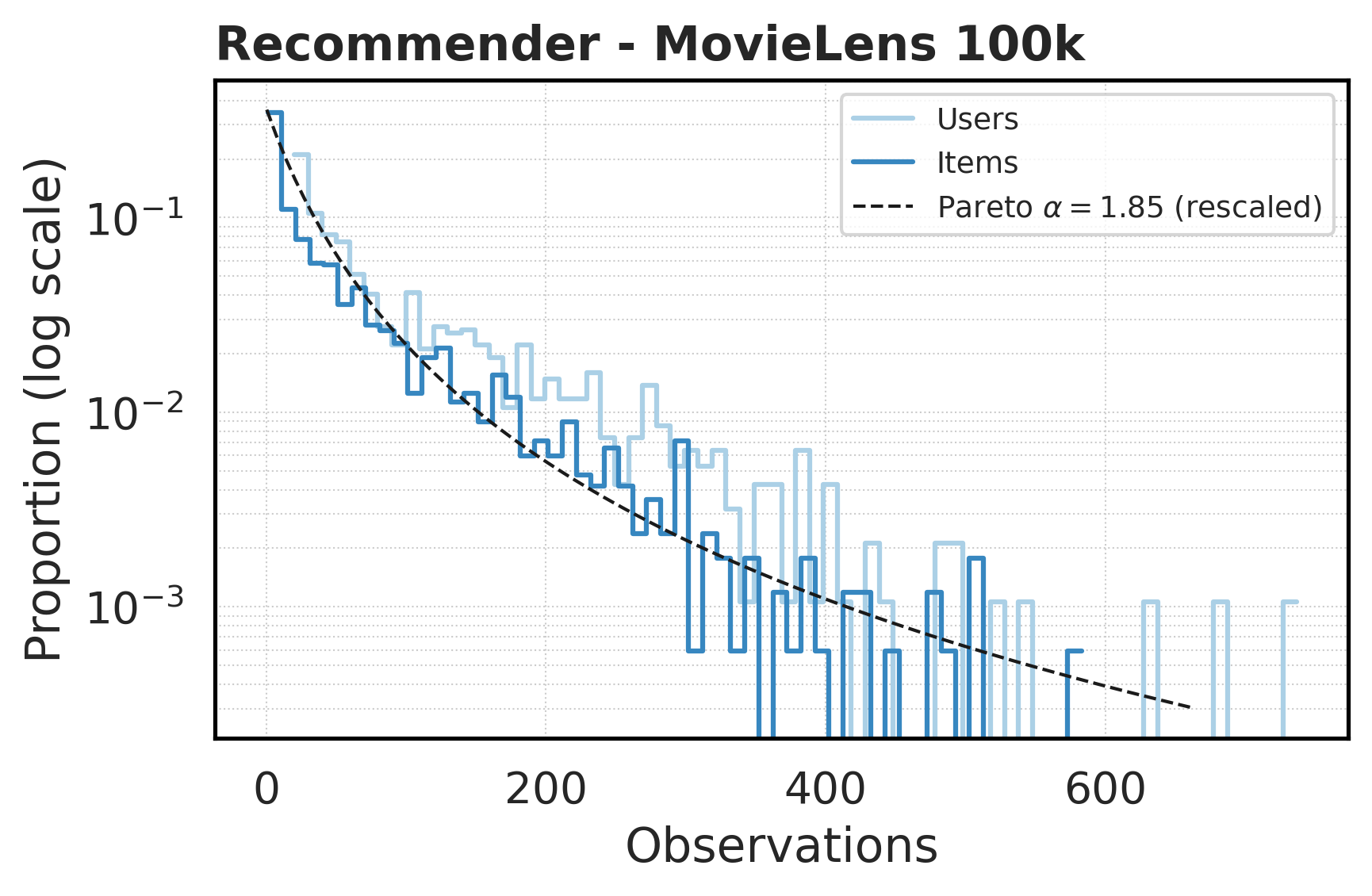}%
        \includegraphics[width=.5\linewidth]{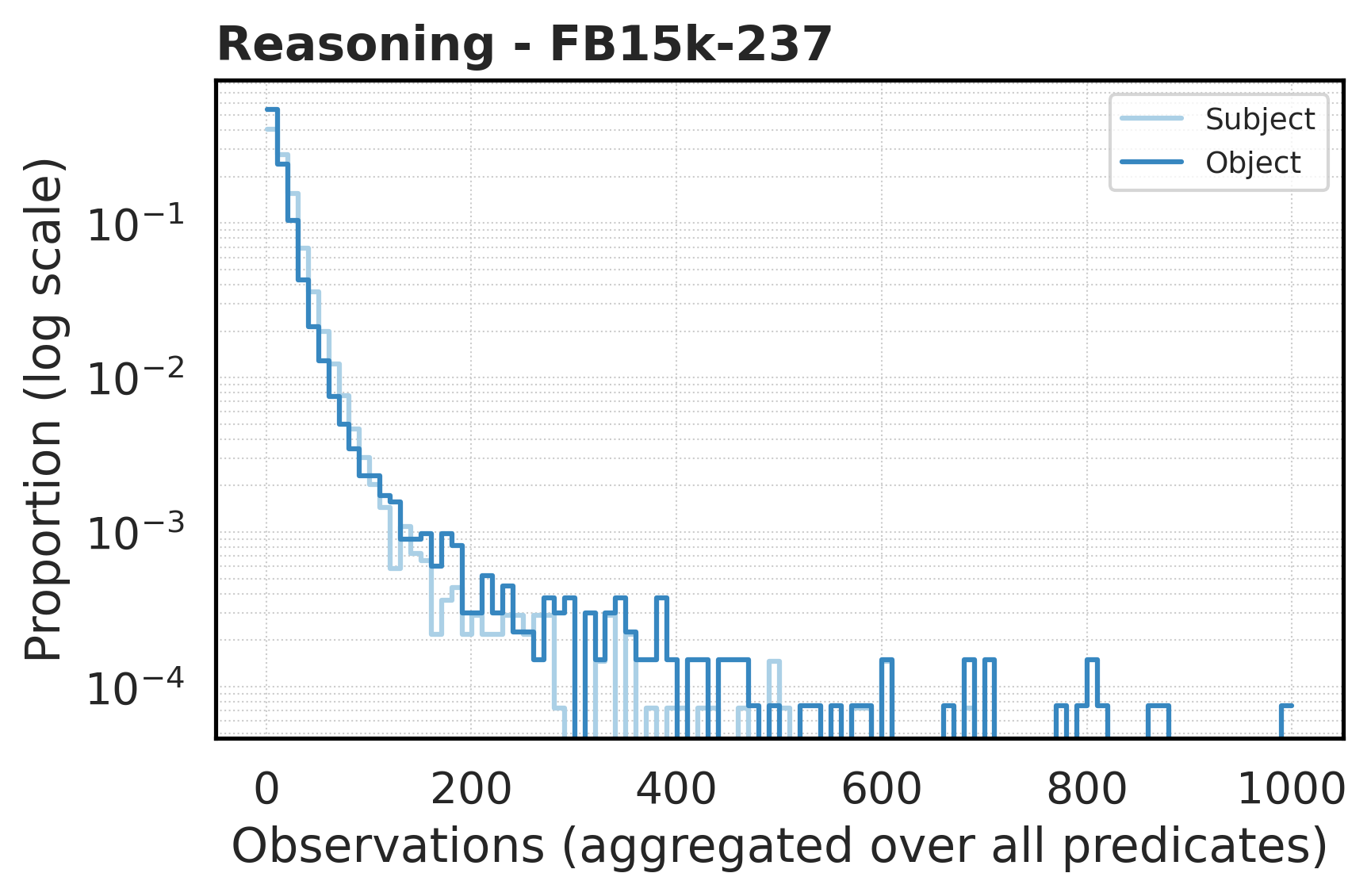} 
        \caption{Heavy-tailed samples}\label{fig:degree-joint}
    \end{subfigure}%
    \begin{subfigure}[T]{.4\linewidth}
        \centering
        \vspace{1ex}
        \resizebox{.95\linewidth}{!}{%
        \begin{tikzpicture}[thick,every node/.style={draw, rounded corners, inner sep=1em},scale=0.9]
        \small
        \draw[dotted] (-4,-1.3) -- (11,-1.3);
        \draw[dotted] (4,1.5) -- (4,-4);
        \node[fill=metabg,align=center] (txt) {``Humans are mortal and \\Socrates is a human"};
        \node[below=5em of txt, align=center, fill=metablue!20] (spo) {\itshape (Socrates, isA, human)\\ \itshape (human, hasProperty, mortal)};
        \node[right=8em of spo, dotted, fill=metablue!20] (reas) {\itshape (Socrates, hasProperty, mortal)};
        \node[above=5.7em of reas, dotted, fill=metabg] (ans) {``Socrates is mortal"};
        \node[above=0.1cm of txt, draw=none] (ol) {\sffamily Observed};
        \node[above=0.4cm of ans, draw=none] (il) {\sffamily Inferred};
        \node[left=of txt, yshift=3em, draw=none,rotate=90,align=center] (nl) {\sffamily Natural\\ \sffamily Language};
        \node[left=of spo, yshift=3em, draw=none,rotate=90] (nl) {\sffamily Symbolic};

        \draw[->] (txt) -- (spo) node[midway,draw=none,fill=white, inner sep=0.25em, align=center] {knowledge representation\\ \textit{(decode)}};
        \draw[->] (spo) -- (reas) node[midway,draw=none,fill=white, inner sep=0.25em] {reasoning};
        \draw[->] (reas) -- (ans) node[midway,draw=none,fill=white, inner sep=0.25em, align=center] {question answering\\ \textit{(encode)}};
    \end{tikzpicture}%
} 
        \vspace{3ex}
        \caption{Reasoning}\label{fig:reasoning}       
    \end{subfigure}
    \caption{%
    (\subref{fig:degree-joint}) \figtitle{Heavy-tailed samples} in recommender and reasoning datasets.
    (\subref{fig:reasoning}) \figtitle{Symbolic reasoning via LLMs.} To validate reasoning capabilities of LLMs, natural language has to be mapped to logical knowledge representations. This shows that validation of reasoning in LLMs is subject to the results of this paper. See also \cref{fig:tensor,app:ternary}.}
\end{figure}
\begin{example}[Recommender Systems]\label{ex:recsys}
Recommender systems are concerned with inferring the true preferences of a user over all items from a set of revealed preferences sampled from \(\sdist\).
As such they are a typical example for \((f, \sdist)\) where 
the target distribution \(\tdist\) corresponds to the uniform distribution over all possible interactions. 
Importantly, \(\sdist\) is typically influenced by social processes and sampling bias as well as heavy-tailed distributions are well documented in recommender systems.
For instance, an important factor for sampling biases are feedback loops, e.g., that past recommendations influence which recommendations are shown in the future \citep{Krauth2022-zc,Chaney2018-vk}. 
Another source of sampling bias is user feedback, which is often biased towards items with high ratings \citep{Steck2010-ty}, as well as popularity bias~\citep{Abdollahpouri2019-te,Papadopoulos2012-dv}. 
Popularity bias leads directly to heavy-tailed distributions in the degree distribution of the sample graph~\citep{Papadopoulos2012-dv,Barabasi1999-ua}. See also \cref{fig:degree-joint} for evidence of this property on \textsc{MovieLens}.
\end{example}

\begin{example}[Symbolic reasoning and QA]\label{ex:reasoning}
Reasoning and question answering over symbolic knowledge representations are another key example for \((f, \sdist)\).
In this setting, factoids are represented in form of \textit{(subject, predicate, object)} triples and the task is to infer the truth value for \emph{any} unknown factoid, i.e., for a uniform target distribution \(\tdist\).
Importantly, while facts about the world itself do not need to be influenced by social processes, our available knowledge about them, i.e., \(\sdist\), often is. In addition to aspects such as popularity bias, causes for this can range from which questions are studied in science~\citep{Kuhn1970-hd,Lacey2005-qs}, over how data is collected~\citep{Jo2020-vt}, to who has access to the internet and the ability to contribute to knowledge~\citep{Wikipedia_contributors_undated-kd}.
Consequently, heavy-tailed distributions are also well-documented in this setting. For instance,~\citet{Steyvers2005-ak} showed that semantic networks typically follow heavy-tailed degree distributions. Similar distributions have been observed in large-scale knowledge graphs such as \textsc{DBPedia}~\citep{Auer2007-qr}, \textsc{Yago}~\citep{Suchanek2007-hw}, \textsc{Freebase}~\citep{Bollacker2008-nj}, and \textsc{Wikidata}~\citep{Vrandecic2014-ih}. See also \cref{fig:degree-joint} for evidence of this property on \textsc{FB15k}.
Importantly, this setting applies to any reasoning task over factoids in general --- irrespective of the data representation. For instance, the validation of reasoning capabilities for general purpose question answering in systems such as \textsc{LLaMA}~\citep{Touvron2023-gx,Dubey2024-al} and \textsc{ChatGPT}~\citep{OpenAI2023-fj} needs to follow this blueprint.
See also \cref{fig:reasoning} for an illustration.
\end{example}

\section{Test validity}\label{sec:validity}
To answer \cref{rq1}, I will focus on the test validity of \emph{inference settings}, i.e., whether task, assumptions, and data allow for \emph{any} valid validations at all.
For this purpose, I will use a deductive approach: model validation is \emph{valid} if it is a logical consequence of its assumptions that the difference between its estimate and the true generalization error is bounded with high probability.
To formalize this, let \({h, f : \Set{X} \to \Set{Y}}\) denote functions that map from sample domain \(\Set{X}\) to target domain \(\Set{Y}\). For clarity, I will assume noise-free \(f\) and \(h\). 
Furthermore, let \({\sset \draw \sdist^m = \{x_i\}_{i=1}^m}\) denote a data set of \(m\) samples drawn from a sampling distribution \(\sdist : \Set{X} \to [0,1]\) and let \({\dset = \{(x, f(x)) : x \in \sset\}}\) denote its supervised extension. For notational convenience, I will also write \(\dset \draw \sdist^m\) when \(f\) is clear from context.
In addition, let \(\A \subseteq \{f \mid f : \Set{X} \to \Set{Y}\}\) be the set of all functions from \(\Set{X}\) to \(\Set{Y}\) that are consistent with some set of assumptions on \(f\) such as being low-rank. 
Next, note that \(\A\) and \(\dset\) then induce a set of possible worlds as follows:
\begin{definition}[Possible worlds]\label{def:pworlds}
    Let \(\A\) be a set of assumptions, \(\dset \subset \Set{X} \times \Set{Y}\) a set of observations, and \(f : \Set{X} \to \Set{Y}\). The set of \emph{possible worlds} \(\pworlds\) is then the set of functions consistent with \(\A\) and \(\dset\), i.e.,
    \setlength{\belowdisplayskip}{3pt}%
    \begin{equation*}
        \pworlds = \{f \mid f \in \A\ \land\ \forall (x, y) \in \dset : f(x) = y \} .
    \end{equation*}
\end{definition}
Furthermore, I will consider an inference setting \(\infset\) to be a set of assumptions \(\A\), a \emph{fixed} dataset \(\dset \draw \sdist^m\), a \emph{target distribution} \(\tdist : \Set{X} \to [0,1]\) for which we want to make inferences, and an assumed \emph{distribution over possible worlds} \(\fdist\). Note that if \(\sdist \neq \tdist\), \(\dset\) can not be an i.i.d.~sample from \(\tdist\). For further details and notation see \cref{app:notation,app:framework}.

Next, let \(X\) be a random variable over \(\Set{X}\) and let \(\ell : \Set{Y} \times \Set{Y} \to \R_+\) be a positive loss function. The \emph{risk} of hypothesis \(h\) with respect to a \emph{single} world \(f\) is then denoted by
\begin{equation*}
    \trisk = \E_{X \draw \tdist}[\ell(h(X), f(X))] .
\end{equation*}
Furthermore, let \(\squal\) denote \emph{any} risk measure of a hypothesis \(h\) on some test set \(\tset\). For instance, \(\smash{\squal}\) could denote the empirical risk or a re-weighted estimator such as the Horvitz-Thompson adjusted empirical risk (see also \cref{tab:estimators} in the supplementary material). Hence, \(\smash{\squal}\) does not only cover the standard Monte-Carlo estimator for the i.i.d.~setting, but also estimators used in counterfactual and causal settings.
To determine the test-validity of an inference setting, I am then interested in bounding difference between the estimated risk (\(\theta\)) and the true risk of h (\(\trisk\)). Importantly, it is necessary to consider the risk of \(h\) relative to the distribution \(\fdist\) over all possible worlds 
since no world \(f \in \pworlds\) can be excluded based on \(\dset\) and \(\A\). Hence, test validity is defined as follows:
\begin{definition}[Test validity]\label{def:test-validity}
    Let \(f \draw \fdist\) denote a distribution over possible worlds \(\pworlds\) and let \(\hyps\) denote a hypothesis class.
    Furthermore, 
    let \(\smash{\trisk}\) denote the risk of hypothesis \(h\) for target distribution \(\tdist\) and possible world \(f\). 
    Let \(\smash{\squal} \in \R_+\) denote any empirical risk measure of \(h\) on a test set.
    Then, \infset~is \((\epsilon,\delta)\)-\emph{test-valid} (\emph{test-invalid}) if \(\smash{\squal}\)'s difference to \(\smash{\trisk}\) can (cannot) be bounded accordingly, i.e.,
    \setlength{\belowdisplayskip}{-2em}%
    \begin{equation*}
        \infset \vDash \begin{cases}
        \ \exists \hyps\ \exists h \in \hyps : \Pr_{f \draw \fdist} ( |\squal - \trisk| \leq \epsilon ) \geq 1 - \delta  & \quad\quad (\epsilon, \delta)\text{-test-validity} \\
        \ \forall \hyps\ \forall h \in \hyps : \Pr_{f \draw \fdist} ( |\squal - \trisk| > \epsilon ) > \delta . & \quad\quad (\epsilon, \delta)\text{-test-invalidity}
        \end{cases}
    \end{equation*}
\end{definition}
The conditions in \cref{def:test-validity} for a valid validation setting are very mild since it requires only a single hypothesis class in which \(\theta\) for a single hypothesis has bounded difference to the true risk with high probability.
Since invalidity follows directly from validity via complement rule and negation, the conditions for a validation setting to be invalid are strong: For \emph{any possible hypothesis class} it has to hold that the difference between \(\theta\) and the true risk of \emph{all hypotheses} can not be bounded with sufficient probability. Importantly, both are statements about an inference setting, i.e., the combination of assumptions, observed data, and target distribution, and not about a specific hypothesis (class). %
Furthermore, note that \cref{def:test-validity} implies realizability with regard to the assumptions: if \(\{f \mid \forall (x,y) \in \dset : f(x) = y\} \cap \A = \emptyset\), an inference setting is test-invalid since \(\Pr(\emptyset) = 0\). However, \cref{def:test-validity} imposes no realizability or any other constraints on \(\hyps\). %

\isneurips{}{\paragraph{Necessary conditions for validity}}
Next, note that \cref{def:test-validity} implies straightforward necessary conditions for test validity: 
\begin{restatable}[Necessary condition for test validity]{corollary}{ttvnecs}\label{cor:ttv-necessary}
    Let \(\infset\) be an inference setting, let \({\ell : \Set{Y} \times \Set{Y} \to \R_+}\) be a positive loss function, and let \(\hyps\) be a hypothesis class. Furthermore, let \(\theta \in \R_+\) be any risk estimate for \(h\). 
    Then, if \infset~is \((\epsilon, \delta)\)-test-valid, it must hold that 
    \begin{equation*}
        \exists \hyps\ \exists h \in \hyps : \PrF (\trisk \leq \epsilon + \theta) \geq 1 - \delta .
     \end{equation*}
\end{restatable}
\begin{prf}
    \Cref{cor:ttv-necessary} follows simply via the monotonicity of probability, i.e., it holds that \(1 - \delta \leq {\Pr_{f \draw \fdist} (|\smash{\squal} - \smash{\trisk}| \leq \epsilon)} \leq {\Pr_{f \draw \fdist} (\smash{\trisk} \leq \epsilon + \smash{\squal})}\). This holds for any risk measure \(\smash{\squal}\), loss \(\ell \in \R_+\) and hypothesis \(h\). See \cref{app:lem:necessary} for proof details.
\end{prf}

\section{Test validity under passive data collection in complex systems}\label{sec:results}

In the following, I will provide an overview of the main results as well as high-level proof sketches. For clarity, I will consider only binary relations \(\Set{X} = \Set{X}_1 \times \Set{X}_2\).
For detailed proofs and discussion, as well as extensions to ternary relations, see \cref{app:thm2}. %
To meet ontological parsimony\footnote{See also \cref{app:iid-valid} for further discussion on the importance of ontological parsimony (minimality).} and get insights into the validity of the train-test paradigm, I will focus on \(\fdist\) being the uniform distribution \(\udist\) and \(\A\) imposing only minimal assumptions on \(f\). %

Next, to derive bounds on the validity of inference settings in complex social systems, I will represent possible worlds \(f\) as partially observed matrices which are constructed as follows:
\begin{definition}[Matrix representation]
For a function \(f : \Set{X}_1 \times \Set{X}_2 \to \Set{Y}\) over \emph{finite} sets of size \(|\Set{X}_1| = n_1\) and \(|\Set{X}_2| = n_2\), its \emph{matrix representation} \(\mF \in \R^{n_1 \times n_2}\) is given via
\(\mF_{ij} = f(x_i, x_j)\) for all \((x_i, x_j) \in \Set{X}_1 \times \Set{X}_2\).\footnote{This is trivially extended to higher arity functions using tensor representations. See also \cref{app:ternary}.} In the following, I will use \(f\) and \(\mF\) interchangeably.
\end{definition}
Using this matrix representation of a system, I will show in \cref{cor:ttv-invalid} that the train-test paradigm is invalid if the rank of \(\ff\), i.e., the complexity of the system, exceeds the \(k\)-connectivity of the sample graph \(\sset\) and if \(f\) is chosen uniformly from \(\pworlds\). Here, \(k\)-connectivity is defined as follows:
\begin{definition}[\(k\)-core and \(k\)-connectivity]\label{def:kcore}
    The \(k\)-core (or core of order \(k\)) of a graph is its maximal subgraph such that all vertices are at least of degree \(k\).\footnote{%
        Note that being in the \(k\)-core of \(\sset\) is a stronger condition than having degree \(k\): A node can be outside the \(k\)-core even with a degree larger than \(k\) if enough of its neighbors are outside the \(k\)-core (see also \cref{fig:2core})
    } A graph is \(k\)-connected \emph{if and only if} every vertex is in a core of order at least \(k\).
\end{definition}

\begin{restatable}[Rank-\(k\) underdetermination]{lemma}{underdet}\label{thm:2}
    Let \(\A = \{ f \mid \rank(\ff) \leq k\}\) and let \(\ell \in \R_+\) be a positive loss function. Then, if \(\sset\) is not \(k\)-connected, \(\pworlds\) forms a non-empty vector space.%
\end{restatable}
\begin{prf}
    Since \(\sset\) is not \(k\)-connected, any \(\ff\) with \(\rank(\ff) = k\) can not be \(\sset\)-isomeric. It then holds via \citep[Lemma 5.1]{Liu2019-hj} that \(\pworlds\), i.e., the set of matrices of rank \(k\) or less that are consistent with \(\dset\), form a non-empty vector space. See \cref{app:thm2} for proof details.
\end{prf}

In the spirit of Occam's razor, higher ranks of \(\ff\) correspond to more complex possible worlds.
\Cref{thm:2} establishes then that if the \(k\)-connectivity of \(\obsvgraph\) does not match the complexity of the system \(\ff\), the observations \(\sset\) do not constrain \(\pworlds\) sufficiently and a randomly chosen possible world can be arbitrarily different on the non-observed entries.
Via \cref{cor:ttv-necessary}, \cref{thm:2} implies then that \(k\)-connectivity is necessary for test validity if \(\ell\) belongs to the broad class of scalar Bregman divergences, i.e., widely used loss functions such as the square loss, the log loss, or the KL-divergence (see also \cref{tab:bregman} in the supplementary material). 
\begin{restatable}[Rank-\(k\) test-invalidity]{lemma}{rankkinvalid}\label{cor:ttv-invalid}
    Let \(\A\) be identical to \cref{thm:2}, let \(\ell\) be a scalar Bregman divergence, let \(\fdist\) be the uniform distribution over \(\pworlds\), and let \(\tdist\) be the uniform distribution over \(\Set{X}\). Furthermore, let \(\smash{\squal} \in \R_+\) be any risk estimator on a test set. Then, if \(\sset\) is not \(k\)-connected, \infset~is test-invalid, i.e., it holds for any \(\epsilon > 0\) that
    \setlength{\belowdisplayskip}{3pt}%
    \begin{equation*} 
       \forall \hyps \ \forall h \in \hyps :
       \Pr_{f \draw \fdist} ( | \smash{\squal} - \smash{\trisk} | \leq \epsilon) = 0 .
    \end{equation*}
\end{restatable}
\begin{prf}
    If \(\obsvgraph\) is not \(k\)-connected, \(\pworlds\) is a vector space according to \cref{thm:2}. \Cref{cor:ttv-invalid} follows then from \cref{cor:ttv-necessary} for uniformly sampled \(f \in \pworlds\) and \(h \in \pworlds\) via a simple volume argument.
    For \(h \not\in \pworlds\), the result follows again from \(\pworlds\) being a vector space via the generalized Pythagorean theorem for Bregman divergences~\citep[Eq. 2.3]{Dhillon2008-lv}. See \cref{app:iid-valid} for proof details.
\end{prf}
The consequences of \cref{cor:ttv-invalid} are non-trivial. Under ontological parsimony, it shows that passive data from complex social systems, i.e., the foundation of basically all large-scale AI tasks, can not be used to validate the quality of models if \(\sdist \neq \tdist\). Clearly, no subset of \(\sset\), e.g., cross-validation, can fulfill this task either. 
Importantly, \cref{cor:ttv-invalid} holds not only for empirical risk, but for \emph{any} estimator on \(\dset\), including counterfactual estimators, i.e., methods which are exactly meant to address \(\sdist \neq \tdist\).
This illustrates that \cref{cor:ttv-invalid} is not simply an out-of-distribution or counterfactual estimation problem. Rather, it is caused by a combination of out-of-distribution (\(\sdist \neq \tdist\)) and insufficient data (\(k\)-connectivity \(<\) \(\rank(f)\)). 
Next, I will connect these results to the main result of this work. 
\begin{theorem}[Test validity in complex social systems]\label{cor:1}
    Let \(\infset\) be identical to \cref{cor:ttv-invalid}.
    Furthermore, let \(\sset \draw \sdist^m\) where \(\sdist\) follows power-law distributions such that the degrees of \(x \in \Set{X}_i\) in the sample graph \(\sset\) are drawn i.i.d.~from a regularly-varying power-law distribution \({\Pr(\deg(x) > k) = u(k)k^{-\alpha_i}}\).
    Furthermore, let \(n_i = |\Set{X}_i|\) be the size of domain \(\Set{X}_i\).
    Then, the number  \(V_i\) of nodes in \(\Set{X}_i\) for which test validity holds decreases with a power-law decay in \(\rank(\ff) = k\), i.e,
    \begin{equation*}
        \E[V_i] \leq n_i u(k)k^{-\alpha_i} .
    \end{equation*}
\end{theorem}
\begin{prf}
    Test validity requires the \(k\)-connectivity of \(\sset\) to be greater or equal to \(\rank(\ff)\) via \cref{thm:2,cor:ttv-invalid}. Hence, only subgraphs where all vertices are at least of degree \(k\) can be valid. \Cref{cor:1} follows then via the expected number of nodes with degree at least \(k\) in \(\Set{X}_i\), i.e, \(\E[V_i] = {\textstyle\sum_{x \in \Set{X}_i}} \Pr(\deg(x) \geq k)\).
\end{prf}
For heavy-tailed distributions, most nodes will be outside the required \kcore~for even moderately complex worlds. Hence, \cref{cor:1} shows that the train-test paradigm cannot be valid under ontological parsimony for the vast majority of nodes in realistic social systems. 
\Cref{tab:scaling} illustrates this using parameters that match the well-known Book Crossing dataset. 
\begin{table}[b]
    \caption{\figtitle{Inefficiency of scaling and benchmarks; validity coverage for the Pareto distribution.}}\label{tab:scaling}
    \resizebox{\linewidth}{!}{%
    \begin{tabular}{cccll}
        \toprule
        & & & \textbf{Scaling} & \textbf{Benchmarks} \\
        \(\alpha\) & \(x_\text{min}\) & \(|\Set{X}|\) & Samples needed to increase \kcore~of random node & Nodes with less than \(100\) observations \\ 
        \cmidrule(r){1-3} \cmidrule(lr){4-4} \cmidrule(l){5-5}
        \(2.5\) & \(5\) & \(10^7\) & \(\E_{i \draw \dist{U}}\left[T_i\right] \geq (|\Set{X}|/2)^{\alpha+1} / (\alpha x^\alpha_\text{min})\quad = \quad 2 \cdot 10^{21}\) & \(\E[N] = |\Set{X}|(1 - (x_\text{min} /x)^\alpha)\quad >\quad 9.9 \cdot 10^6\) \\
        \midrule
        & & & \multicolumn{2}{l}{\textbf{Book Crossing}~\citep{Ziegler2005-bc}}\\
        \(\alpha\) & \(x_\text{min}\) & \(|\Set{X}|\) & \multicolumn{2}{l}{Fraction of users with large enough degrees such that train-test measures and inferences are valid}\\
        \cmidrule(r){1-3} \cmidrule(l){4-5}
        \(2.38\) & \(8\) & \(10^5\)& \multicolumn{2}{l}{Rank 8: \(100\%\),\quad Rank 10: \(58.8\%\),\quad Rank 20: \(11.3\%\),\quad Rank 100: \(0.2\%\)}\\
        \bottomrule 
    \end{tabular}}
\end{table}

An immediate next question is then if the issues raised by \cref{cor:1} can simply be solved by scaling, i.e., by collecting more data from \(\sdist\) --- or via manually constructed benchmarks such as BigBench~\citep{Srivastava2023-dy} to extrapolate from their results to the risk on \(\tdist\).
\Cref{cor:inefficiency} answers both questions via \cref{thm:2} (see \cref{app:scaling} for a detailed discussion and proof):~(i) For \emph{scaling}, we can ask how many draws from \(\sdist\) would be necessary such that all nodes are within the \(k\)-core of \(\sset\) with high probability, i.e., how many samples are needed until arriving at a valid test setting.
While there exists no easily computable solution to this problem, we can compute a (weak) lower bound by asking how many samples from \(\sdist\) are needed to sample a random node in \(\Set{X}_i\) \emph{once}.
(ii) For \emph{benchmarks}, we can ask how many nodes would need \emph{at least one} additional data point to arrive at a valid test setting, i.e., how much manual data collection is at least needed to create a benchmark that extrapolates to \(\tdist\).
\begin{corollary}[Inefficiency of scaling and benchmarks]\label{cor:inefficiency}
    Let \infset~and \(\sdist\) be identical to \cref{cor:1}. Furthermore, let (i) \(T_i\) denote the expected number of samples from \(\sdist\) until node \(x_i \in \Set{X}\) is sampled, and let (ii) \(N_j\) denote the number of nodes in \(\Set{X}_j\) with less then \(k\) samples. Then, \(T_i\) scales at least \emph{polynomially} and \(N_j\) scales lineraly in \emph{the size of the domain} \(|\Set{X}|\). Specifically,
    \begin{equation*} 
        \E_{i \draw \dist{U}\{1, |\Set{X}|\}} [T_i] \geq (|\Set{X}| / 2)^{\alpha+1} / (\alpha x_\text{min}^\alpha), \quad \text{and} \quad
        \E[N_j] = |\Set{X}|(1 - (x_\text{min}/ x)^\alpha)
    \end{equation*} 
\end{corollary}
Clearly, sampling from \(\sdist\) is highly inefficient to overcome the issues raised by \cref{cor:1} since (i) it is extremely difficult to get successful samples from the heavy tail (rare events) and (ii) covering all nodes outside sufficiently large {\kcore}s in selective benchmarks is prohibitively expensive. See also \cref{tab:scaling} for examples of these aspects for typical distributions in complex social systems.

\begin{figure}[t]
    \vspace{-1ex}
    \begin{subfigure}[t]{.31\linewidth}
        \centering
        \includegraphics[width=.95\linewidth]{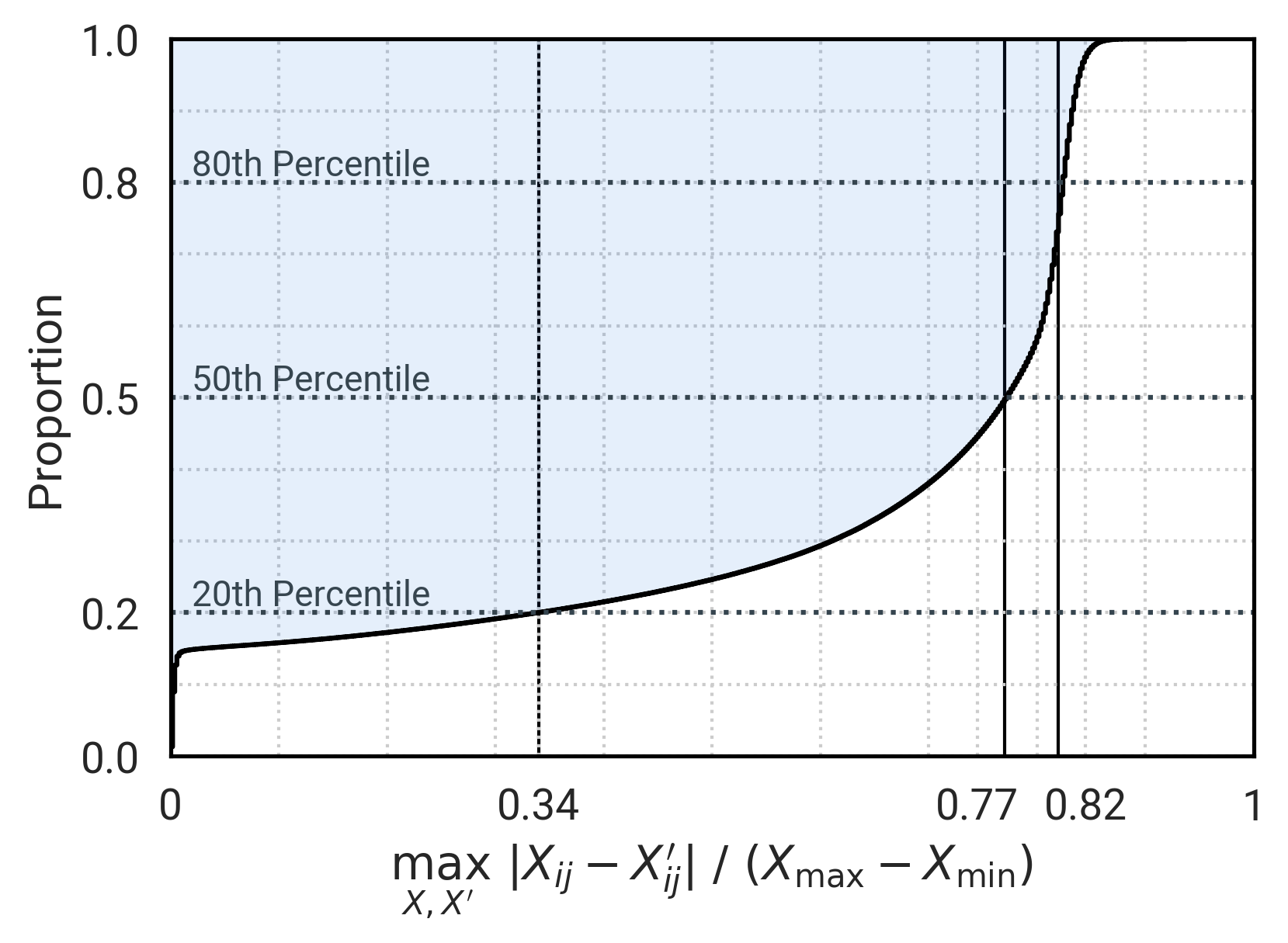}
        \caption{eCDF of Maximum NAE\label{fig:exfit}}
    \end{subfigure}%
    \begin{subfigure}[t]{.31\linewidth}
        \centering
        \includegraphics[width=.95\linewidth]{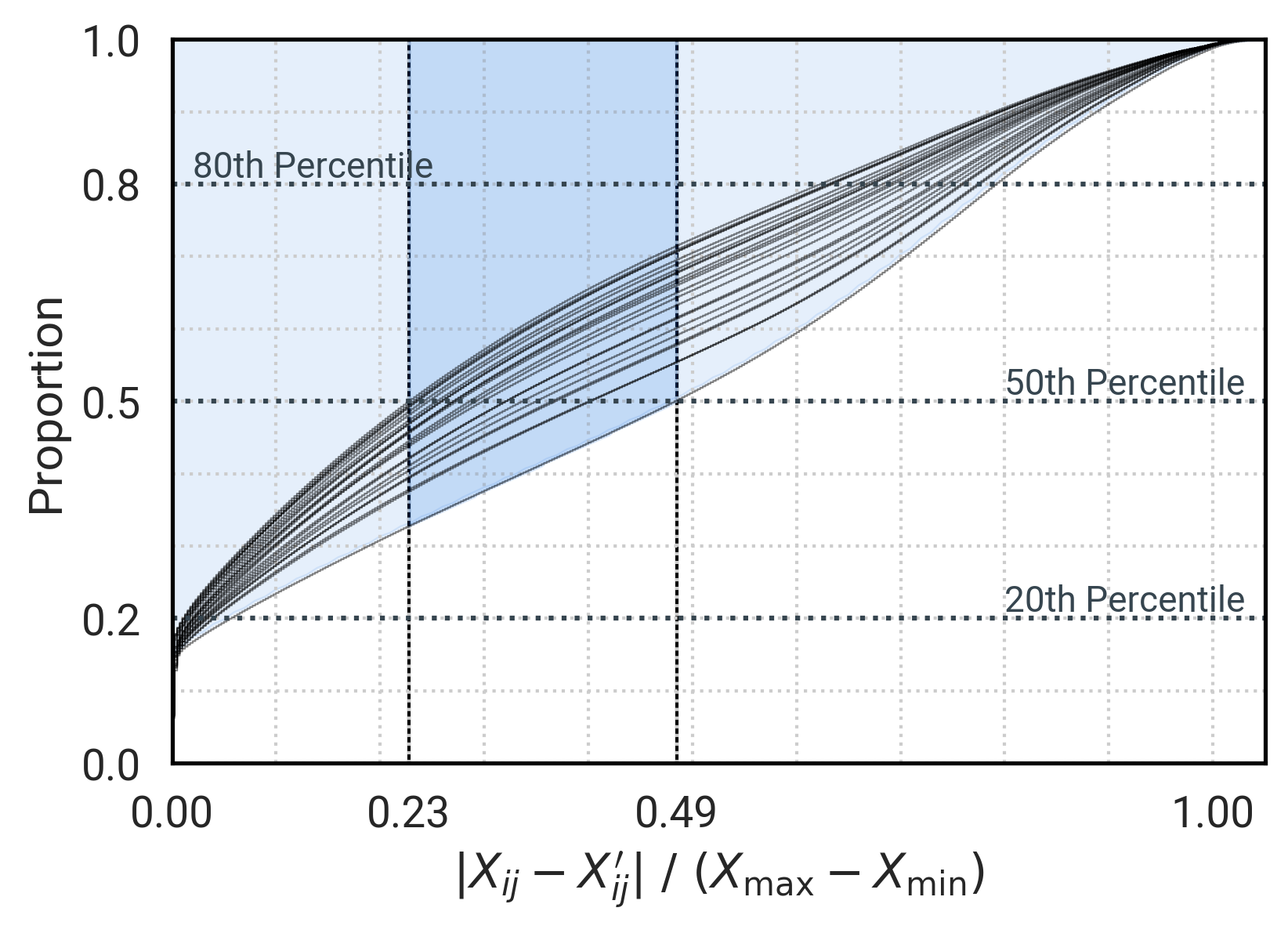}
        \caption{eCDF of Pairwise NAE\label{fig:ecdf}}
    \end{subfigure}%
    \begin{subfigure}[t]{.36\linewidth}
        \centering
        \includegraphics[width=.96\linewidth]{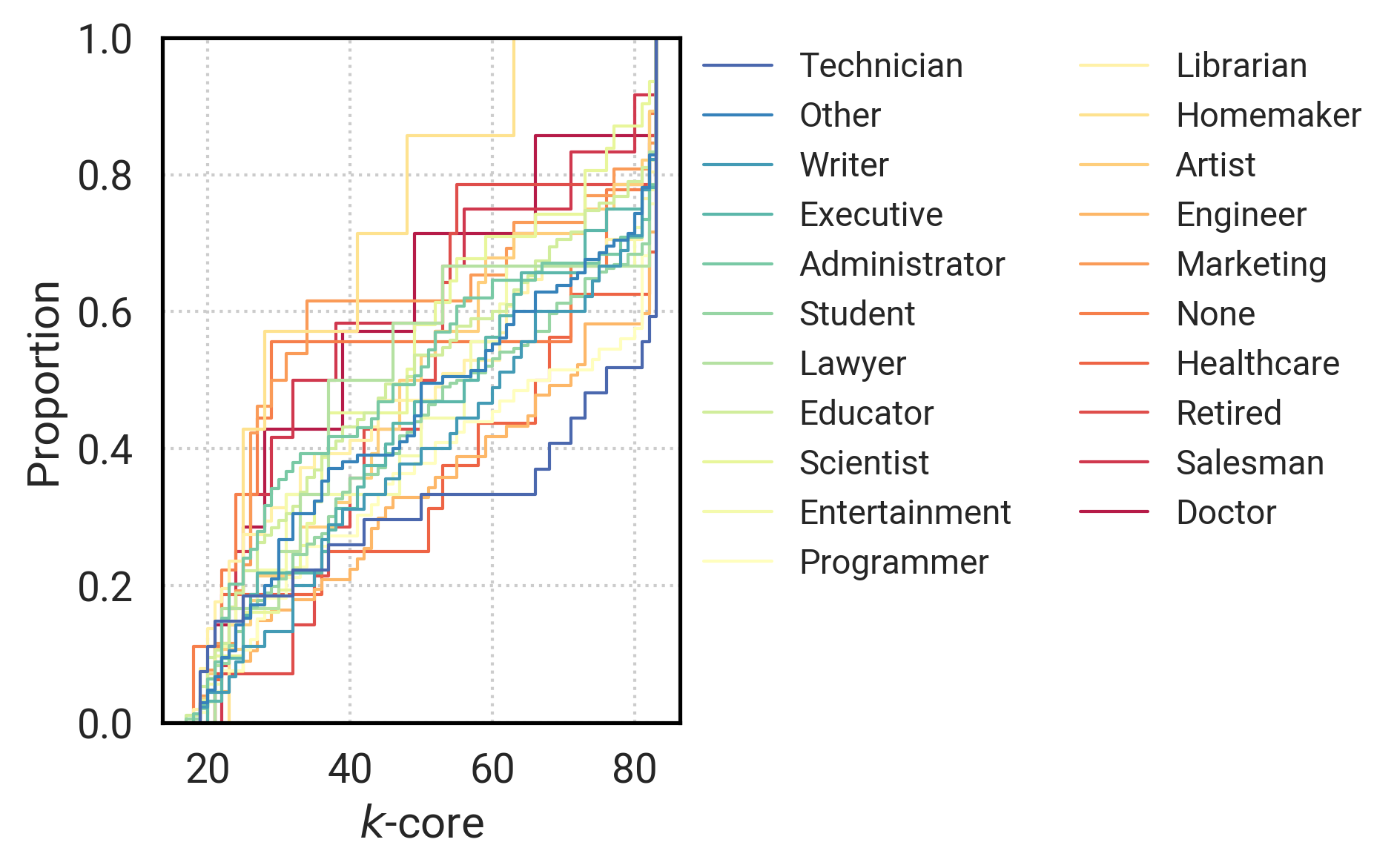}
        \caption{\(k\)-core per occupation\label{fig:ecdf-kcore}}
    \end{subfigure}

    \begin{subfigure}[t]{\linewidth}
        \centering
        \includegraphics[width=.99\linewidth]{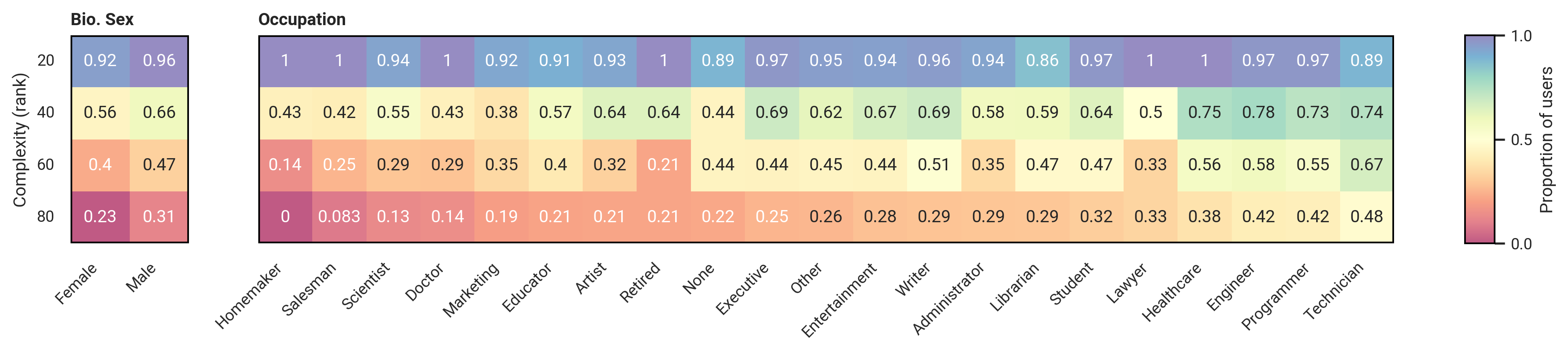}
        \caption{Test-validity per demographic group and model complexity\label{fig:demo}}
    \end{subfigure}
    \caption{\figtitle{\textsc{MovieLens} 100k experiments} 
    (\subref{fig:exfit}) Empirical CDF (eCDF) of maximum NAE over possible worlds. Area over the curve (expected error) shaded.\ %
    (\subref{fig:ecdf}) eCDF of NAE for pairs of possible worlds.\ %
    (\subref{fig:ecdf-kcore}) eCDF \kcore per demographic group.\ %
    (\subref{fig:demo}) Proportion of users for which test-validity holds relative to the rank of \(\ff\).
    }\label{fig:experiments}
    \vspace{-1em}
\end{figure}

\paragraph{\normalfont\bfseries Experimental evidence}\label{sec:experiments}
To illustrate the real consequences of the previous theoretical results, I will now provide experimental evidence based on the \textsc{MovieLens} 100k dataset~\citep{Harper2015-po}, a critical benchmark that has, for years, been widely-used in recommender systems research.
As predicted by \cref{cor:ttv-invalid}, I will show that there exist possible worlds of low complexity that all explain the observed data equally well but are widely different on the unobserved data. Hence, \emph{any} quality metric that is inferred on this benchmark, or subsets of it, can not be informative about the true generalization error. 
For this purpose, I fit \(p=100\) matrices of rank \(k=50\) to the observed data \(\dset\).
All matrices, or possible worlds, fit the observed data and rank constraint with error below \(10^{-3}\) and \(10^{-2}\), respectively. 
See \cref{app:method-possworlds} for details.
For a pair of possible worlds \((\ff, \ff')\) , I compute then the normalized absolute error (NAE) for each \emph{unobserved} entry \((i,j) \not \in \obsvgraph\) via
\(\text{NAE}(\ff_{ij}, \ff'_{ij}) = |\ff_{ij} - \ff'_{ij}|/(\ff_{\text{max}} - \ff_{\text{min}})\). This informs us about how different pairs of possible worlds can be on the unobserved data.
\Cref{fig:exfit,fig:ecdf} shows the empirical CDF (eCDF) of the NAE over unobserved entries for such pairwise comparisons of possible worlds as well as the worst-case over all worlds per entry. 
From \cref{fig:exfit}, it can be seen that the worst case error across possible worlds per entry is substantial for the vast majority of unobserved entries. For instance, for 50\% of entries the NAE is above 77\% of the worst case error.
For arbitrary pairs of possible worlds, the situation is similar, where, depending on the particular pair of worlds, the NAE is between 23\% to 49\% for 50\% of entries.
Furthermore, the area over the eCDF curves in \cref{fig:ecdf} corresponds directly to the risk for a pair of possible worlds and is again substantial for all pairs (see \cref{app:auccdf} for details). Since any possible world can be the ``true'' world this shows again that the test error for any subset of this benchmark can not be informative for the true generalization error of this task.

In addition to the NAE, \cref{fig:ecdf-kcore} shows the cumulative distribution of users within cores of order \(k\) per demographic group for \textsc{MovieLens} 100k. It can be seen that the cumulative distribution can vary significantly between different demographics. 
For instance, while only 25\% of ``homemakers'' are in a \(k\)-core larger than 50, 40\% of ``technicians'' are in a \(k\)-core larger than 80. 
It follows from \cref{cor:ttv-invalid}, that test-validity will therefore also vary significantly between demographic groups (if we assume that there are no significant differences in the complexity of preferences between groups). 
\Cref{fig:demo} illustrates this point by showing the proportion of users for which test-validity holds relative to the rank of a model. It can be seen that there exist clear differences already for moderately complex worlds. For instance, for a model of rank 60, test-validity would hold for 67\% of ``technicians'' while it would only hold for 14\% of ``homemakers''. 
Clearly, this has important implications for fairness, bias, and whether recommender systems \emph{work for everyone}.  

\section{Related work}\label{sec:related}
The no-free-lunch theorems for machine learning~\citep{Wolpert1996-hh,Sterkenburg2021-ou} share important similarities to this work as both consider the expected risk over possible worlds.
However, the results in this paper are stronger and directly applicable to current machine learning practice. While the NFL theorems consider the performance over all possible worlds without any restrictions --- an assumption that is too restrictive in most instances --- the results of this paper show that even for relatively strong assumptions about the set of possible worlds, e.g., low-rank structures, valid model validation is not generally possible for passive data collection in complex social systems. %
In motivation, this paper is also related to the works \citep{DAmour2022-xw,Black2022-ct,Fisher2018-hd,Semenova2022-ka,Marx2019-to,Watson-Daniels2023-id} which study outcomes of underspecification in ML pipelines, model multiplicity and Rashomon sets. In the restricted context of personalized prediction, \citet{Monteiro-Paes2022-xf}, discusses related limits to testing and estimation. \citet{Schaeffer2023-ro} discuss whether seemingly emergent capabilities of LLMs are rather a result of insufficient metrics.
In statistics,~\citet{Meng2018-km} analyzed a scaling-related question similar to this paper: Given a carefully collected survey with low response rate (small data) or a large, self-reported dataset without data curation (big data), which dataset should one trust more to estimate population averages? 
Outside machine learning, validity theory has a long history in fields such as psychology and sociology. Here, test validity is considered a measure of the degree to which a test measures what it is intended to measure~\citep{Cronbach1955-jo} and has been studied extensively in the context of psychological tests~\citep{Messick1989-wd} and educational testing~\citep{Kane2013-ih}. Increasingly, these notions of validity, have also been considered in machine learning~\citep{Coston2022-df,Recht2022-qr,Raji2022-px,Abebe2022-jk}.
 
With regard to technical tools, this paper is also closely related to prior work in matrix completion. For instance, \citep{Kiraly2015-qt} studied the problem of unique and finite completability of matrices and derived similar \(k\)-core related bounds using determinantal varieties and algebraic geometry.
\citet{Srebro2010-ob} studied the problem of matrix completion based on non-uniform samples such as power-laws but assume that \(\sdist = \tdist\). \citet{Meka2009-fo} focused on power-law samples for \(\sdist \neq \tdist\) and, consistent with this work, require at least \(k\) samples per row and column to guarantee completability of a rank-\(k\) matrix. \citet{Cheng2018-pb} derive similar results based on graph \(k\)-connectivity. Related to non-i.i.d.~observations,~\citep{Liu2019-hj} developed a framework to provide necessary conditions for matrix completion under deterministic sampling. \Cref{thm:2} is based on these results. 
Different to these prior works, I provide formal impossibility results for test validity based on passive data in complex social systems. This allows to gain rigorous insights into the epistemic limits of what we can know based on this form of data collection. See also \cref{app:related} for further related work.

\section{Discussion}\label{sec:discussion}
The results in this paper provide new insights into the validity of the train-test paradigm when data is passively collected from complex social systems. 
In particular, I have shown that there exists \emph{no free delivery service} of data that allows for test validity on a global scale in this setting. While valid inferences are possible with respect to the sampling distribution \(\sdist\) and within high \(k\)-cores, they are unlikely if \(\tdist\) extends to the entirety of the system. Hence, test validity depends on the interplay between task (\(\tdist\)), the complexity of the system (\(\A\)), and the \(k\)-connectivity of the sample graph (\(\obsvgraph\)) underlying the observed data (\(\dset\)), what is a \emph{combinatorial} property of the data. 
These results are attained by establishing novel \emph{necessary conditions} for which validation is possible. As AI systems are increasingly applied in conditions for which sufficient conditions of validity are difficult to guarantee, understanding such minimal conditions can provide guidelines into developing better and more robust systems.
Importantly, it can help to demarcate inference goals that are not meaningful from ones that are attainable. It helps to understand the limits of what we can know and which questions are futile to ask. This work provides a first step in this direction by establishing such epistemic limits of AI in complex social systems.

Furthermore, I have shown that the sub-system for which valid inferences are possible shrinks rapidly with the complexity of the system and that a na{\"\i}ve application of the scaling paradigm is prohibitively inefficient to  overcome these validity issues.
As a consequence, solving many complex AI tasks are unlikely to come for free through scaling or for cheap through extrapolating from limited small-scale benchmarks. Instead, there exists an inherent trade-off between data quality, quantity, and task complexity. If we want to avoid asking AI systems to solve simpler tasks (e.g., non-out-of-distribution or smaller scope), new data curation efforts are likely needed.
Due to the substantial amount of data that would have to be collected, centralized data collection is often infeasible to overcome the validity issues of this paper. Instead, decentralized methods such as \emph{participatory data curation} could provide a way forward. This aligns with insights from fairness which also highlight the need for participatory methods in data collection~\citep{Jo2020-vt}. Similar arguments apply to the importance of open science and open-source models in this context.

Importantly, the theoretical results of this paper also provide direct insights into how to improve data collection for model validation via its \(k\)-core conditions. In particular, \cref{thm:2} and \cref{cor:inefficiency} imply two clear objectives for targeted data collection: 
(a) collecting data points that increase the \(k\)-connectivity of the sample graph and
(b) collecting data points that increase the size of the \(\rank(f)\)-core of the sample graph, where \(\rank(f)\) is the complexity of the world that we want to assume.
Pursuing (a) would increase the complexity of the world that can be assumed such that model validation is still valid for the entire sample graph,
while pursuing (b) would increase the size of the subgraph for which a \(\rank(f) = k\) assumption would still yield valid model validation.
Hence, both objectives are based on the k-core conditions of this work and can be computed from a given sample graph. 
Creating new mechanisms for efficient data collection based on these insights is therefore a very promising avenue for future work.

\section*{Acknowledgments}
I gratefully acknowledge the valuable feedback from Léon Bottou, Smitha Milli, Tina Eliassi-Rad, Mark Tygert, and anonymous reviewers which all helped to improve various versions of this paper. 

\printbibliography

\clearpage
\newpage
\beginappendix

\section{Notation}\label{app:notation}
Random variables are denoted by italic uppercase letters, e.g., \(L, S, X\).
Sets are denoted by calligraphic uppercase letters, e.g., \(\Set{X}, \Set{S}\).
Constants are indicated with lowercase greek letters, e.g., \(\epsilon, \rho\). 
Functions and scalar are denoted by lowercase letters, e.g., \(f, g, h\) and \(x, y\).
Matrices and higher-order tensors are indicated with bold uppercase letters, e.g., \(\mat{F}, \mat{U}\).

\begin{table}[h]
    \centering
    \caption{Notation}
    \begin{tabular}{ll}
        \toprule
        \textbf{Concept} & \textbf{Notation} \\
        \midrule
        Possible world & \(f : \Set{X} \to \Set{Y}\) \\
        Hypothesis & \(h : \Set{X} \to \Set{Y}\) \\
        Loss & \(\ell : \Set{Y} \times \Set{Y} \to \R_+\) \\
        Sample distribution & \(\sdist : \Set{X} \to [0,1]\) \\
        Target distribution & \(\tdist : \Set{X} \to [0,1]\) \\
        Social system & \((f, \sdist)\) \\
        Sample graph & \(\obsvgraph \draw \sdist^m\) \\
        Test set & \(\dset = \{(x, f(x)) : x \in \obsvgraph \}\) \\
        Risk & \(\trisk = \E_{X \draw \tdist}[\ell(h(X), f(X))]\) \\
        Estimated risk & \(\squal \in \R_+\) \\
        \bottomrule
    \end{tabular}
\end{table}

\section{Validity framework}\label{app:framework}
In this work, I am interested in the validity of \emph{inference settings}, i.e., whether assumptions and observations allow for \emph{any} valid inferences at all.
To formalize this, I will take the following high-level approach:
\begin{description}
    \item[Inference setting] An inference setting consists of a set of assumptions \(\A\), a \emph{fixed} dataset \(\dset\) which is collected from a sampling distribution \(\sdist\), and a target distribution \(\tdist\) for which we want to make inferences. Note that \(\sdist\) is not guaranteed to be identical to \(\tdist\). Hence, we're concerned with out-of-distribution generalization settings.
    \item[Expected risk over possible worlds] Assumptions \(\A\) and observed data \(\dset\) define a set of possible worlds \(\pworlds\) that is consistent with \(\A\) and \(\dset\). Given a probability distribution \(\fdist\) over \(\pworlds\), I am then interested in the expected risk over all possible worlds that are consistent with \(\A\) and \(\dset\).
    \item[Validity] An inference setting is valid, if the expected risk over possible worlds can be bounded meaningfully at all, i.e., if there exists \emph{at least one} hypothesis class for which the generalization error of at least \emph{a single} hypothesis can be bounded sufficiently. 
\end{description}

To approach the question of validity, learning theory has traditionally focused nearly exclusively on \emph{sufficient conditions} for valid inferences. Under active data collection, i.e., in scenarios where one can control exactly how data is collected, sufficient conditions are highly attractive since they provide exact specifications for inferences to be valid with high probability. 
However, under passive data collection, the situation is reversed. Sufficient conditions for the validity of inferences usually place highly restrictive demands on the data collection process (e.g., i.i.d.~samples or simple random sampling) which are challenging to satisfy even when data is collected carefully in an active way. Since passive data collection, by definition, exerts no control over the sample generating process, these sufficient conditions are not met with near certainty. 
For this reason, I am focusing here on \emph{necessary conditions} for validity, i.e., conditions that must always be satisfied for inferences to be valid. Under passive data collection, necessary conditions can provide important insights since they need to hold for any data collection process or, conversely, can be used to identify scenarios where inferences are not valid with high probability.

\begin{table}[b]
    \caption{\figtitle{Estimators and quality measures.}}\label{tab:estimators}
    \resizebox{\linewidth}{!}{%
    \begin{tabular}{lcccc}
        \toprule
        & \multicolumn{2}{c}{Estimators} & \multicolumn{2}{c}{Risk Measures} \\
        \cmidrule(r){2-3} \cmidrule(l){4-5}
        & \textbf{Monte Carlo} & \textbf{Horvitz-Thompson} & \textbf{Empirical Risk} & \textbf{HT Weighted Emp. Risk} \\ 
        \midrule
        Estimator from \(\sset \draw \sdist^m\)
        & \(\frac{1}{m}\sum_{x \in \sset} x\) 
        & \(\frac{1}{m}\sum_{x \in \sset} \frac{x_i}{p_{\tdist}(x)}\) 
        & \(\frac{1}{m}\sum_{x \in \sset} \ell(h(x), f(x))\)
        & \(\frac{1}{m} \sum_{x \in \sset} \frac{\ell\left(h(x), f(x)\right)}{p_{\tdist}(x)}\) \\
        \cmidrule(r){1-1}\cmidrule(lr){2-2}\cmidrule(lr){3-3}\cmidrule(lr){4-4}\cmidrule(l){5-5}
        Estimated expected value
        & \(\E_{X \draw \sdist}[X]\) 
        & \(\E_{X \draw \tdist}[X]\) 
        & \(\E_{X \draw \sdist}[\ell(h(X), f(X))]\)
        & \(\E_{X \draw \tdist}[\ell(h(X), f(X))]\) \\
        \bottomrule
    \end{tabular}}
\end{table}

\subsection{Connection to No-Free-Lunch theorems}
The validity framework of \cref{sec:validity} and the No-Free-Lunch theorems are closely connected.
First, consider the \emph{expected risk over all possible worlds} relative to \(\fdist\), i.e., 
\begin{equation}\label{eq:expected-risk}
    \E_{f \draw \fdist}\left[ \trisk \right] = \E_{f \draw \fdist}\E_{X \draw \tdist}[\ell(h(X), f(X))] .
\end{equation}
\Cref{eq:expected-risk} is then akin to the objectives considered in the seminal \emph{No Free Lunch} (NFL) theorems~\citep{Wolpert1996-hh,Sterkenburg2021-ou}.
For instance, the NFL theorem for supervised learning can be written as \(\forall \Lambda : \E_{f \draw \dist{U}} \E_{X \draw \tdist}[\ell(h_{\Lambda(\dset)}(X), f(X))] = 1/2\), where \(\dist{U}\) is the uniform distribution over all possible worlds in an assumption-free setting (i.e., \(\A = \emptyset\)), \(\ell\) is the \(0/1\)-loss, and \(h_{\Lambda(\dset)}\) is the hypothesis derived from a finite sample \(\dset\) with algorithm \(\Lambda\). 
In contrast to the NFL theorems --- where \(\A = \emptyset\) implies an induction-hostile universe --- my focus is on induction-friendly settings (\(\A \neq \emptyset\)) but where \(\dset\) is sampled from a complex social system.
Since \(\trisk\) is a non-negative random variable, we can then connect \cref{def:test-validity} and \cref{eq:expected-risk} via upper and lower bounds based on Markov's inequality.
\begin{definition}[Markov's inequality] 
    Let \(X\) be a non-negative random variable and \(a > 0\). Then 
    \[
        \Pr(X \geq a) \leq \E[X]/a .
    \]
\end{definition}
Hence, it follows that the expected risk over all possible worlds is large for invalid settings since it holds that
\[
    \E_{f \draw \fdist}[\trisk] \geq \epsilon \cdot \PrF (\trisk > \epsilon)
\]

\subsection{Importance of ontological parsimony and test validity in the i.i.d.~setting}\label{app:iid-valid}
The strong appeal of the train-test paradigm is that, with careful data collection, we require no further ontological assumptions to ensure the validity of the model validation procedure.
In particular, if we have a test set that is sampled independently from \(\tdist\), it follows straightforwardly from Hoeffding's inequality that we can meaningfully bound the approximation error over this test set~\citep[Theorem 11.1]{Shalev-Shwartz2014-jv}. Let \(\tset \draw \tdist^m\) be a test set of size \(m\), sampled i.i.d. from the target distribution \(\tdist\). Then, it holds that
\begin{equation*} 
    \Pr_{\tset \draw \tdist^m} \left( \left| L^{\tset}_{hf} - L^{\tdist}_{hf} \right| \leq \sqrt{\frac{\log (2/\delta)}{2m}}\right) \geq 1 - \delta .
\end{equation*}
Importantly, this holds for \emph{any} hypothesis \(h\), \emph{any} algorithm \(\Delta\), and \emph{any} training set \(\dset\).
Hence, under careful data collection where we know that if the test set is sampled i.i.d.~from \(\tdist\), \emph{any hypothesis can be validated based on the observed data only}.

This property, i.e., that we can evaluate the performance of a model without further assumptions on the model itself, is crucial to compare the performance of different methods since different architecture, inference, and hyperparameter choices correspond to different assumptions.
Maybe more importantly, this property is also crucial to validate our model assumptions on observed data (given that the sampling assumption holds), since otherwise we could only make statements relative to that our model assumptions hold, which is, of course, much weaker and not informative.
Hence, if we need to make model specific assumptions for the validation error to be informative for the generalization error, the train-test paradigm would be relatively meaningless.
Of course, the (considerable) challenge is to collect \(m\) i.i.d.~samples from the \emph{true target} distribution \(\tdist\) which can not be guaranteed and is an important assumption on the data collection process.

\subsection{Ternary and higher arity relations}\label{app:ternary}
\begin{figure}
    \centering
    \begin{subfigure}[t]{.3\linewidth}
        \centering
        \resizebox{0.9\linewidth}{!}{\raisebox{2em}{%
        \begin{tikzpicture}[thick,scale=2]
        \draw[pattern=north east lines, pattern color=Grey!50, draw=black] (0,0) ellipse (2.5 and 1.5);
        \draw[fill=white] (1,0.4) ellipse (1 and 0.8);
        \draw[fill=white] (-1,-0.1) ellipse (0.6 and 1.1);

        \node at (0,1.7) {Assumptions \(\A\)};
        \node[align=center] at (1,0.3) {Possible worlds\\ \(\pworlds\)};
        \node[align=center] at (-1,-0.1) {Hypotheses\\ \(\hyps\)};

        \node[circle,fill=black, inner sep=2pt] (fstar) at (0,0.4) {};
        \node[circle,fill=black, inner sep=2pt] (f) at (0.8,0.9) {};
        \node[circle,fill=black, inner sep=2pt] (h) at (-0.8,0.4) {};
        \node at (-0.95,0.5) {\(h\)};
        \node at (0.2,0.3) {\(f^\star\)};
        \node at (0.95,0.9) {\(f\)};
        \draw[-] (fstar) -- (h);
        \draw[-] (f) -- (h);
        \draw[-,dotted] (fstar) -- (f);
    \end{tikzpicture}%
}}
        \caption{Validity relationships}\label{fig:framework}
    \end{subfigure}%
    \begin{subfigure}[t]{.3\linewidth}
        \centering
        \resizebox{0.7\linewidth}{!}{%
    \begin{tikzpicture}[auto matrix/.style={matrix of nodes,
  draw,thick,inner sep=0pt,
  nodes in empty cells,column sep=-0.2pt,row sep=-0.2pt,
  cells={nodes={minimum width=1.9em,minimum height=1.9em,
   draw,very thin,anchor=center,fill=white,
   execute at begin node={%
   $\vphantom{x_|}\ifnum\the\pgfmatrixcurrentrow<4
     \ifnum\the\pgfmatrixcurrentcolumn<4
      #1_{\the\pgfmatrixcurrentrow \the\pgfmatrixcurrentcolumn}
     \else 
      \ifnum\the\pgfmatrixcurrentcolumn=5
       #1_{\the\pgfmatrixcurrentrow n}
      \fi
     \fi
    \else
     \ifnum\the\pgfmatrixcurrentrow=5
      \ifnum\the\pgfmatrixcurrentcolumn<4
       #1_{m \the\pgfmatrixcurrentcolumn}
      \else
       \ifnum\the\pgfmatrixcurrentcolumn=5
        #1_{mn}
       \fi 
      \fi
     \fi
    \fi  
    \ifnum\the\pgfmatrixcurrentrow\the\pgfmatrixcurrentcolumn=14
     \cdots
    \fi
    \ifnum\the\pgfmatrixcurrentrow\the\pgfmatrixcurrentcolumn=41
     \vdots
    \fi
    \ifnum\the\pgfmatrixcurrentrow\the\pgfmatrixcurrentcolumn=44
     \ddots
    \fi$
    }
  }}}]
 \matrix[auto matrix=Y^3,xshift=3em,yshift=3em,ampersand replacement=\&](matz){
  \& \& \& \& \\
  \& \& \& \& \\
  \& \& \& \& \\
  \& \& \& \& \\
  \& \& \& \& \\
 };
 \matrix[auto matrix=Y^2,xshift=1.5em,yshift=1.5em,ampersand replacement=\&](maty){
  \& \& \& \& \\
  \& \& \& \& \\
  \& \& \& \& \\
  \& \& \& \& \\
  \& \& \& \& \\
 };
 \matrix[auto matrix=Y^1,ampersand replacement=\&](matx){
  \& \& \& \& \\
  \& \& \& \& \\
  \& \& \& \& \\
  \& \& \& \& \\
  \& \& \& \& \\
 };
 \draw[thick,-stealth] ([xshift=1ex]matx.south east) -- ([xshift=1ex]matz.south east)
  node[midway,below,rotate=45] {\(\Set{X}_3\)};
 \draw[thick,-stealth] ([yshift=-1ex]matx.south west) -- 
  ([yshift=-1ex]matx.south east) node[midway,below] {\(\Set{X}_2\)};
 \draw[thick,-stealth] ([xshift=-1ex]matx.north west)
   -- ([xshift=-1ex]matx.south west) node[midway,above,rotate=90] {\(\Set{X}_1\)};
\end{tikzpicture}%
}
        \caption{Tensor representation}\label{fig:tensor}
    \end{subfigure}%
    \begin{subfigure}[t]{0.4\linewidth}
        \centering
        \resizebox{0.9\columnwidth}{!}{\raisebox{4em}{%
    \begin{tikzpicture}[thick,node distance=1cm,scale=0.8]
    \foreach \pos/\name in {{(0,0)/a}, {(1,0)/b}, {(2,1)/c}, {(3,0)/d}, {(2,-1)/e}, {(-1,1)/f}, {(-1,-1)/g}, {(-2,1)/h}, {(-2,-1)/i}, {(-3,0)/j}}{
        \node[circle,draw=black,fill=metafg,minimum size=0.25cm] (\name) at \pos {};
    }
    \foreach \pos\name in {{(0,1)/k}, {(0,-1)/l}, {(-4,1)/m}, {(-4,-1)/n}, {(-5,0)/o}, {(-5,-1)/p}, {(4,0)/q}, {(4,1)/r}, {(4,-1)/s}, {(5,0)/t}, {(5,1)/u}, {(5,-1)/v}} {
        \node[circle,draw=black,fill=white,minimum size=0.25cm] (\name) at \pos {};
    }
    \foreach \source/\target in {a/b, b/c, b/e, e/d, c/d, a/f, a/g, f/h, g/i, f/i, g/h, h/j, i/j}{
        \draw (\source) -- (\target);
    }
    \foreach \source/\target in {a/k, c/l, j/m, j/n, n/o, n/p, c/r, r/q, q/s, q/t, r/u, s/v} {
        \draw[dotted] (\source) -- (\target);
    }
\end{tikzpicture}%
%
}}
        \caption{2-core of a graph}\label{fig:2core}%
    \end{subfigure}
    \caption{%
        (\subref{fig:framework}) \figtitle{Relation between assumptions, possible worlds, and hypotheses} Assumptions \(\A\) define a set of functions \(f\) of which a subset are possible worlds \(\pworlds\), i.e., those functions which are also consistent with observations \(\dset\). While hypotheses \(\hyps\) will often be equivalent to \(\A\), they can also be a proper subset of \(\A\) and do not need to overlap with \(\pworlds\), e.g., due to additional assumptions or computational requirements that constrain \(\hyps\). The functions \(f\),\(f^\star\), and \(h\) indicate the relevant objects for the necessary conditions in \cref{cor:ttv-necessary} as well as their relationships (solid and dotted lines).
        (\subref{fig:tensor}) \figtitle{Tensor representation of a function} \(f : \Set{X}_1 \times \Set{X}_2 \times \Set{X}_3 \to \R\).
        (\subref{fig:2core}) \figtitle{Illustration of the \(2\)-core of a graph}. Nodes within the \(2\)-core are indicated by black. Nodes outside the \(2\)-core are indicated as white, edges that are removed when reducing to the \(2\)-core are indicated as dotted.}\label{fig:tensor-2core}
\end{figure}
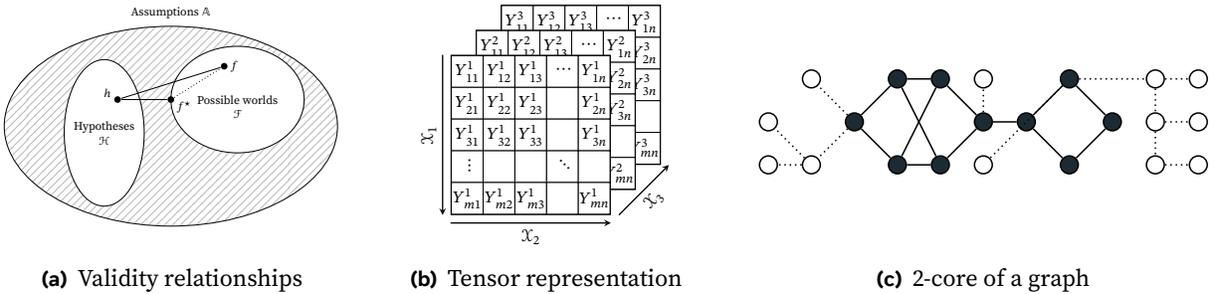

First, note that higher arity functions can be represented as tensors of the same order as follows (see also \cref{fig:tensor} for a visualization):
\begin{definition}[Tensor representation]\label{def:tenrep}
    For a function \(f : \Set{X}_1 \times \Set{X}_2 \times \cdots \times \Set{X}_k \to \R\) over \emph{finite} sets of size \(|\Set{X}_i| = m_i\), we can construct its \emph{tensor representation} \(\mF \in \R^{m_1 \times m_2 \times \cdots \times m_k}\) via
    \(\mF_{ij \ldots, k} = f(x_i, x_j, \ldots, x_k)\) for all \(x_i \in \Set{X}_1, x_j \in \Set{X}_2, \ldots, x_k \in \Set{X}_k\).
\end{definition}

A trivial extension of \cref{cor:1} and its related results can then be obtained by considering the rank of the projection of the tensor representation of \(f\) onto its matrix presentation such that \({\mF_{ij} = \sum_k f(i, j, k)}\) (what equals the assumption that the rank of the predicate mode in \(f\) is one). Since predicates in many knowledge graphs are very sparse, the sum over \(k\) will often preserve this sparsity and the associated heavy-tailed distributions. In this case, the results of the matrix case extend directly to the tensor case. If this is not the case, it is necessary to extend the matrix analysis of \cref{thm:2} to the tensor case and consider cases where the predicate mode can have rank larger than one. However, this is beyond the scope of this paper and reserved for future work.

\section{Complex social systems}\label{app:complex}
\begin{figure}[t]
    \begin{subfigure}[t]{.225\linewidth}
        \centering
        \raisebox{1.3em}{%
        \resizebox{.95\linewidth}{!}{%
    \begin{tikzpicture}[thick]
    \node[draw,circle, minimum width=3.5em] (Xij) {$f_{ij}$};
    \node[draw,circle, minimum width=3.5em, right=3em of Xij] (Sij) {$S_{ij}^t$};
    \node[right=1.5em of Xij,yshift=-3em,font=\footnotesize] (it) {$t = 1, \ldots, T$};
    \node[right=1.3em of Xij,yshift=-5.5em,font=\footnotesize] (ix) {$i,j = 1, \ldots, M$};
    \node[draw,fit=(Xij) (ix.east),rectangle, rounded corners, inner sep=1em] {};
    \node[draw,fit=(Sij) (it),rectangle, rounded corners] {};
    \draw[->] (Xij) -- (Sij);
    \draw (Xij) edge[loop above] node {$ij \neq k\ell$} (Xij);
    \draw (Sij) edge[loop above] node {$t' \rightarrow t$} (Sij);
\end{tikzpicture}%
}}
        \caption{Sampling bias}\label{fig:prop-bias}    
    \end{subfigure}%
    \begin{subfigure}[t]{.23\linewidth}
        \centering
        \raisebox{.5em}{%
        \resizebox{.95\linewidth}{!}{%
    \begin{tikzpicture}
    \begin{axis}[domain=1:5,samples=100,xmin=1,xmax=5,ymin=0,ymax=3,grid,thick,scale=0.7,cycle list name=linestyles]
    \addlegendimage{empty legend};
    \addlegendentry{$\alpha$}
    \pgfplotsinvokeforeach{1,1.5,2,2.5,3}{
        \addplot {#1 / (x^(#1 + 1))};
        \addlegendentry{#1};
    }
    \end{axis}
\end{tikzpicture}%
}}
        \caption{Pareto distribution}\label{fig:prop-powerlaw}
    \end{subfigure}%
    \begin{subfigure}[t]{.525\linewidth}
        \centering
        \includegraphics[width=\linewidth]{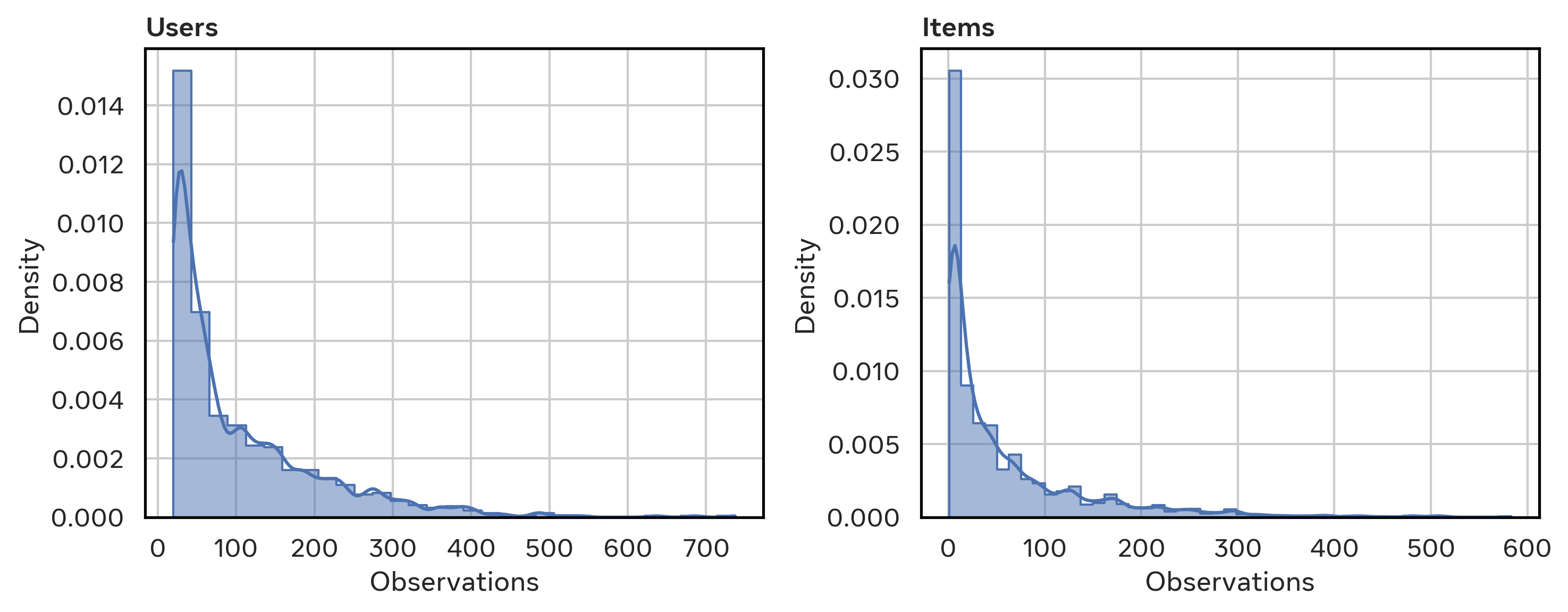}
        \caption{Degree distribution of \textsc{MovieLens} 100k}\label{fig:movielens-dist}
    \end{subfigure}
    \caption{\figtitle{Properties of complex social systems} (\subref{fig:prop-bias}) Graphical model of sampling bias.\ (\subref{fig:prop-powerlaw}) Illustration of power-law distribution on the example of the Pareto distribution.\ (\subref{fig:movielens-dist}) Degree distribution of \textsc{MovieLens} 100k.}
\end{figure}
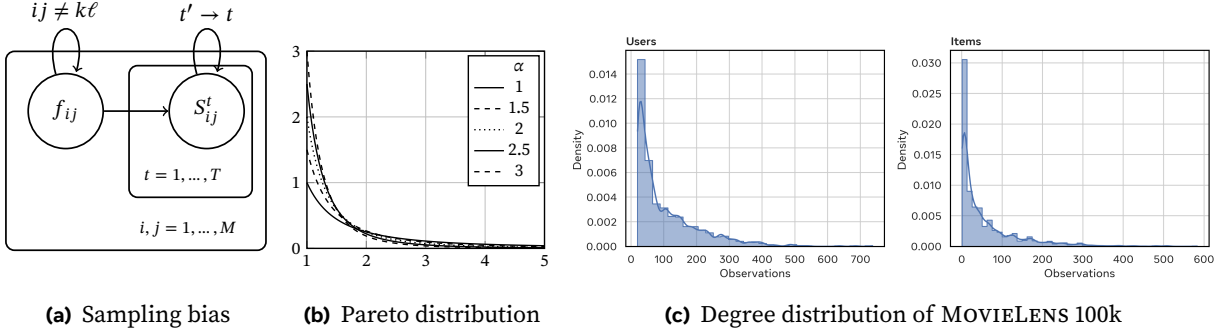

In the following, I will discuss how sampling bias and heavy-tailed distributions can occur, and can be connected, in complex social systems.
First, \emph{sampling bias} is concerned with how \(\obsvgraph\) is collected. Most standard inference methods assume i.i.d. samples from \(\tdist\), but it is well know that this assumption can be easily violated when sampling in complex systems.
\begin{definition}[Sampling bias]\label{def:bias}
Let \(\obsvgraph_{ij}^t\) denote the random variable corresponding to entities \((i,j) \in \Set{X}_1 \times \Set{X}_2\) \(j\) being samples at time \(t\). Samples in complex social systems can then neither be assumed to be independent across time nor independent with regard to the target value \(f_{ij}\), i.e.,
\begin{equation*}
P(S_{ij}^t | S_{ij}^{t - s}) \neq P(S_{ij}^t)
\quad \text{and} \quad
P(S_{ij}^t | f_{ij}) \neq P(S_{ij}^t) .
\end{equation*}
Higher arity relations are defined analogously. See also \cref{fig:prop-bias} for the assumed sample dependencies.
\end{definition}
Sample biases as in \cref{def:bias} can be causes by aspects such popularity bias, i.e., if popular items are more likely to be sampled, and quality biases, i.e., if items with higher values for \(f_{ij}\) are more likely to be sampled.

A prime example of how sampling bias in the form of popularity bias can lead to power-law distributions, is the influential Barabasi-Albert model~\citep{Barabasi1999-ua}. In this model of complex networks, nodes are added to a network one by one and are connected to existing nodes with a probability proportional to their degree, i.e., popularity. Formally, this model is defined as follows:
\begin{definition}[Barabasi-Albert model]\label{def:bamodel}
    Let \(G = (\Set{X}, \Set{E})\) be a graph. Furthermore, let \(\Pr(i \draw_t j)\) denote the probability that the edge \(i \draw j\) is added at time \(t\) to \(\Set{E}\) and let \(\kappa_i\) denote the degree of node \(x_i\). The Barabasi-Albert model generates then a graph \(G\) as follows: 
    \begin{enumerate} 
        \item Start with a small connected graph \(G_0\) with \(m\) nodes. 
        \item At each time step \(t > 0\), add a new node \(x\) to \(G\) and connect it to \(m\) existing nodes in \(G\) with a probability proportional to their degree, i.e.,
        \[
            \Pr(i \draw_t v) = \frac{\kappa_i}{\sum_{j \in \Set{X}} \kappa_j}
        \]
     \end{enumerate}
\end{definition}
It is then well known that \cref{def:bamodel} leads to a power-law degree distribution in \(G\), i.e., a distribution where the probability of a node having \(k\) connections is proportional to \(k^{-\alpha}\) for some \(\alpha > 0\).

\begin{figure}[tb]
    \centering
    \begin{subfigure}{.5\textwidth}
        \centering
        \includegraphics[width=0.95\textwidth]{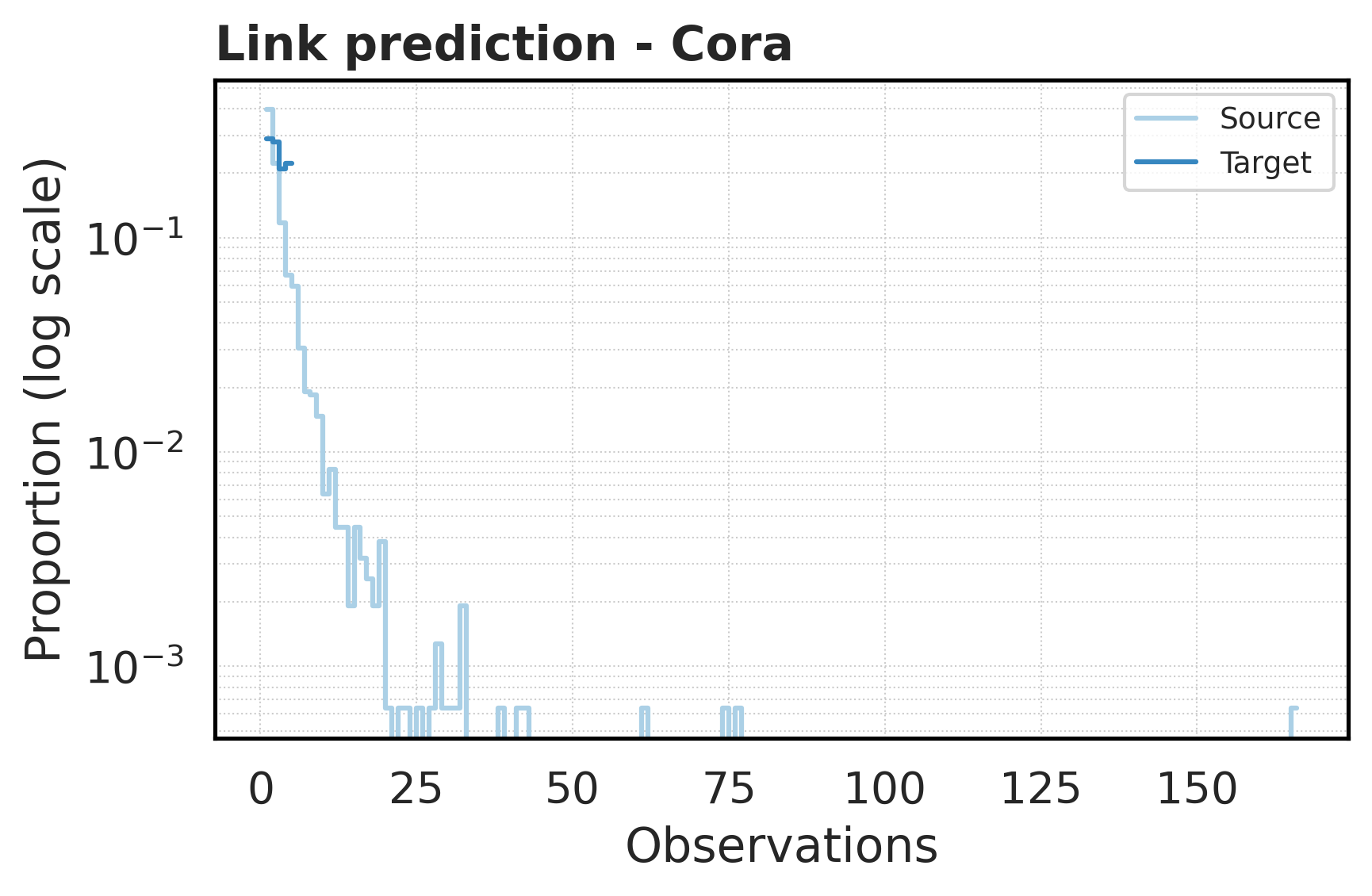}
        \caption{}\label{fig:cora}
    \end{subfigure}%
    \begin{subfigure}{.5\textwidth}
        \centering
        \includegraphics[width=0.95\textwidth]{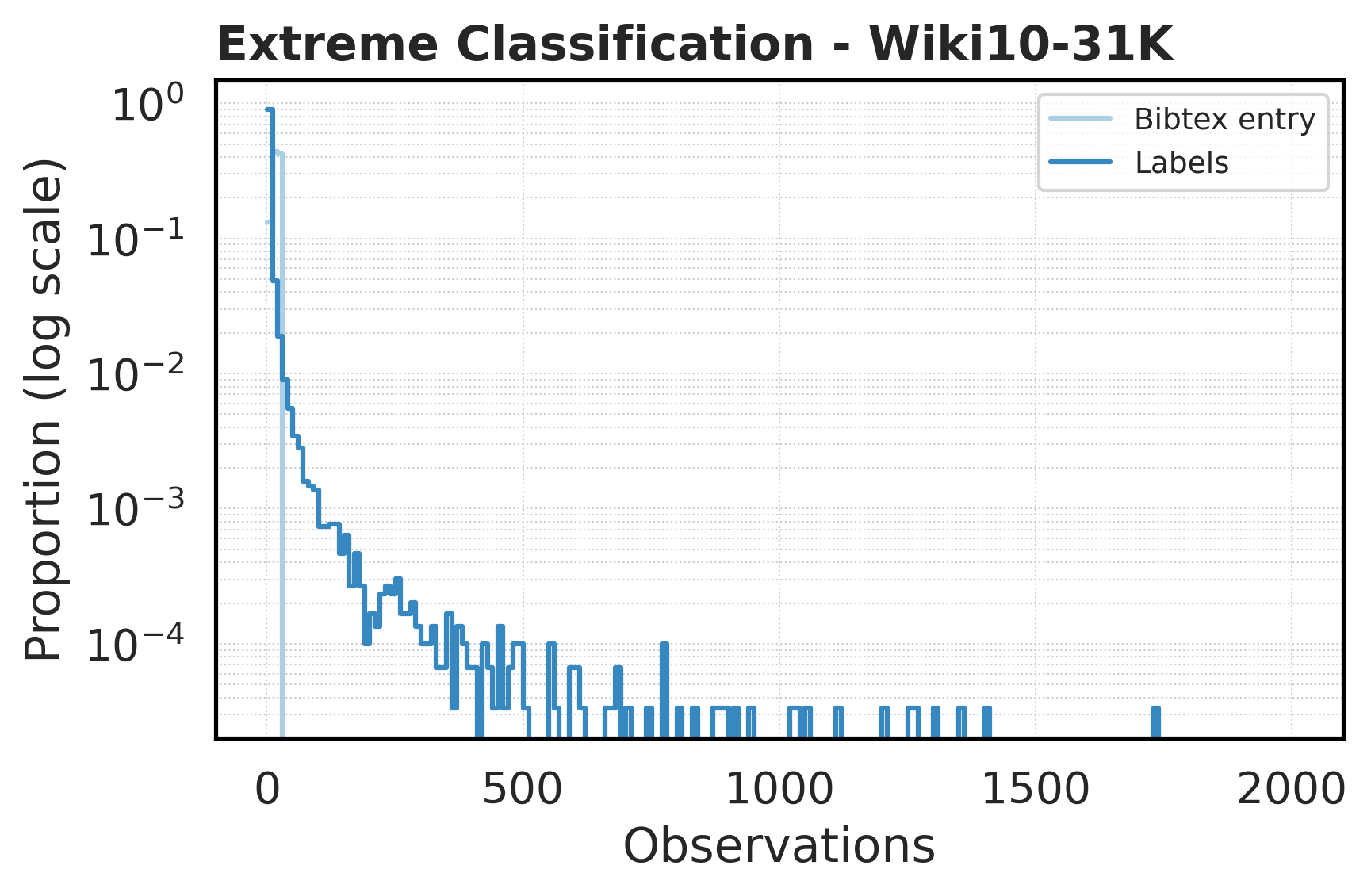}
        \caption{}\label{fig:wiki10}
    \end{subfigure}
    \caption{\textbf{Sample graph degree distributions} for widely used benchmark datasets. (a) Graph learning and link prediction via Cora (b) Extreme classification via Wiki10-31k. As can be seen, all benchmarks follow similar heavy-tailed distributions in their sample graph as the MovieLens dataset in the main text. As such, these benchmarks are subject to the same results and pathologies.}
\end{figure}

\section{Proof \cref{cor:ttv-necessary} (Necessary condition for test validity)}\label{app:lem:necessary}

\ttvnecs*

\begin{proof}
    First, note that \(|\squal - \trisk| \leq \epsilon\) is equivalent to \(\squal - \trisk \leq \epsilon \land \trisk - \squal \leq \epsilon\). Furthermore, we have
    \begin{equation*}
        \{f \mid \squal - \trisk \leq \epsilon \land \trisk - \squal \leq \epsilon\} 
        \subseteq
        \{f \mid \trisk - \squal \leq \epsilon\} . \label{eq:app:invalid}
    \end{equation*}
    It follows then simply from the monotonicity of probability 
    that
    \begin{equation*}
        1 - \delta \leq \PrF(|\squal - \trisk| \leq \epsilon) \leq \PrF(\trisk - \squal \leq \epsilon) = \PrF(\trisk \leq \epsilon + \squal). \qedhere
    \end{equation*}
\end{proof}

\section{Proof \cref{thm:2} (Rank-\(k\) underdetermination)}\label{app:thm2}
I will first introduce the concept of \obsvgraph-isomerism and connect it to \(k\)-connectivity. I will then use these results to proof \cref{thm:2}.

First, let \(\sset_{i, \cdot} = \{j : (i,j) \in \sset\}\) denote the
set of observed columns for row \(i\) and \(\sset_{\cdot, j} = \{i : (i,j) \in \sset\}\) denote the observed rows for column \(j\). 
Let \(\sset_{i, \cdot}[\mF] \in \R^{m \times |\sset_{i, \cdot}|}\) be the sub-matrix of \(\mF \in \R^{m \times n}\) which is obtained by restricting the columns of \(\mF\) to the indices in \(\sset_{i,\cdot}\).
Similarly, let \(\sset_{\cdot, j}[\mF] \in \R^{|\sset{\cdot, j}| \times n}\) be the sub-matrix of \(\mF \in \R^{m \times n}\) which is obtained by restricting the rows of \(\mF\) to the indices in \(\sset_{\cdot, j}\).
Then, \(\obsvgraph\)-isomerism is defined as follows:
\begin{definition}[\obsvgraph-Isomeric]\label{def:isomeric}
    Let \(\mF \in \R^{m \times n}\) and let \(\obsvgraph \subseteq \{1,\ldots, m\} \times \{1,\ldots,n\}\) with \(\sset_{\cdot, j} \neq \emptyset\). Then, \(\mF\) is called \(\obsvgraph\)-isomeric iff
    \begin{align*}
        \rank\left(\sset_{i, \cdot}[\mF]\right) & = \rank(\mF),\quad \forall i \in 1,\ldots , m \quad \text{and}\\
        \rank\left(\sset_{\cdot, j}[\mF]\right) & = \rank(\mF),\quad \forall j \in 1,\ldots , n.
    \end{align*}
\end{definition}

\begin{corollary}[Necessary condition for \(\sset\)-isomerism]\label{cor:isomeric}
    Let \(\sset\) be a sample graph and let \(\rank(\mF) = k\). If \(\mF\) is \(\sset\)-isomeric, then it must hold that \(\sset\) is \(k\)-connected. 
\end{corollary}
\begin{proof}
    First, note that \(\rank(\mX) \leq \min(m, n)\) for any \(\mX \in \R^{m \times n}\). Hence, it follows from \cref{def:isomeric} that for a rank-\(k\) matrix to be \(\sset\)-isomeric, each column and row needs to have at least \(k\) observed entries. Since this is equivalent to \(k\)-connectivity, the result follows.
\end{proof}

\underdet*
\begin{proof}
    Since \(\rank(f) = k\) and \(\sset\) is not \(k\)-connected, it follows from \cref{cor:isomeric} that \(f\) is not \(\sset\)-isomeric.
    Hence, it holds via \citep[Theorem 3.2]{Liu2019-hj} that there exist infinitely many matrices \(\ff'\) that all explain the observed data \(\sset\) perfectly, i.e.,
    \begin{equation*}
        \ff' \neq \ff
        \quad \land \quad
        \rank(\ff') \leq \rank(\ff)
        \quad \land \quad
        \ff'_{ij} = \ff_{ij} \quad \forall (i,j) \in \sset.
    \end{equation*}
    Moreover, it follows from \citep[Lemma 5.1]{Liu2019-hj} that this set of possible worlds \(\pworlds\) forms a non-empty vector space \(\Set{V}\).
\end{proof}

\section{Proof \cref{cor:ttv-invalid} (Rank-\(k\) test invalidity)}\label{app:ttv-invalid}
To prove \cref{cor:ttv-invalid}, I will first show the following auxiliary proposition:

\begin{table}[b]
    \centering
    \small
    \caption{Examples of Scalar Bregman Divergences}\label{tab:bregman}
    \begin{tabular}{llll}
        \toprule
        \textbf{Divergence} & \(\ell(x, y)\) & \textbf{Divergence} & \(\ell(x, y)\) \\
        \cmidrule(r){1-2}\cmidrule(l){3-4}
        Squared loss & \((x - y)^2\) &
        KL-divergence & \(x \log(x/y)\) \\
        Log loss & \(x \log(x / y) + (1-x) \log((1-x)/(1-y))\) &
        Itakura-Saito & \(\frac{x}{y} - \log(x/y) - 1\) \\
        \bottomrule
    \end{tabular}
\end{table}

\begin{proposition}[Risk inequality for Bregman projection]\label{prop:pythagorean}
    Let \(\ell : \Set{Y} \times \Set{Y} \to \R_+\) be a scalar Bregman divergence and let \(f^\star = \arginf_{f} \smash{\trisk}\) be the Bregman projection of \(h\) onto a vector space \(\Set{F}\). Then, it holds that
    \begin{equation*}
        L^{\tdist}_{ff^\star} \leq \trisk . \label{eq:app:pythagorean}
    \end{equation*}
\end{proposition}
\begin{proof}
    First, note that \(\trisk\) is simply a convex combination of scalar Bregman divergences, i.e.,
    \begin{equation*}
        \trisk = \sum_{x \in \Set{X}} \ell(f(x), h(x))p_T(x) .  
    \end{equation*}
    Hence, \(\trisk\) itself is a (separable) Bregman divergence. \Cref{eq:app:pythagorean} follows than from the generalized Pythagorean theorem for Bregman divergences~\citep[Eq. 2.3]{Dhillon2008-lv} since every vector space is a convex set and \(f^\star\) is the projection of \(h\) onto \(\Set{F}\), i.e., it holds that
    \begin{equation*}
        \trisk \geq L^{\tdist}_{ff^\star} + L^{\tdist}_{f^\star h} \geq L^{\tdist}_{ff^\star} . \qedhere
    \end{equation*}
\end{proof}

\rankkinvalid*
\begin{proof}
Since the standard uniform distribution is not defined on an entire vector space, I will instead consider the limit of the class of uniform distributions of balls of radius \(r\). %
Next, since \(\ell\) is Borel measurable, \({\mathcal{B}(\ff,\epsilon)} = \{f' \mid L^{\tdist}_{\ff \ff'} < \epsilon\}\) defines a measurable set around each \(\ff \in \pworlds\).
Furthermore, let \(\text{Vol}\Set{B}(\ff,\epsilon)\) denote the volume of such an \(\epsilon\)-``ball''. Via the change of variables formula, we know then that the volume of \(\Set{B}(f, r \cdot \epsilon)\), i.e., the volume of the original ball stretched in all directions by \(r \geq 1\) is given by 
\begin{equation*}
    \text{Vol}\ \mathcal{B}(\ff, r \cdot \epsilon) = 
    \int_{r \cdot \epsilon} dx = \int_\epsilon r^{\text{dim}\Set{V}} dy = r^{\text{dim}\Set{V}} \cdot \text{Vol} \Set{B}(f,\epsilon).
\end{equation*}
Next, let \(\dist{U}_r\) denote the uniform distribution over \(\Set{B}(\ff, r \cdot \epsilon)\). Then, the probability of sampling a point inside \({\mathcal{B}(\ff,\epsilon)}\) when drawing points uniformly from \(\mathcal{B}(\ff, r \cdot \epsilon)\) with \(r \geq 1\) is given by
\begin{equation*}
    \forall \ff \in \pworlds : 
    \Pr_{\ff' \draw \dist{U}_r} \left(L^{\tdist}_{\ff\ff'} \leq \epsilon \right) 
    = 
    \frac{\text{Vol}\ \mathcal{B}(\ff, \epsilon)}{\text{Vol}\ \mathcal{B}(\ff, r \cdot \epsilon)} 
    = \frac{1}{r^{\text{dim}\Set{V}}} .
\end{equation*} 
Moreover, if \(\Set{V}\) is non-empty it follows that \(\text{dim}\Set{V} \geq 1\). Hence, as we increase \(r\) to span large parts of \(\Set{V}\), it holds that
\begin{equation}
    \forall \epsilon \forall \ff \in \pworlds : 
    \lim_{r \to \infty} 
    \Pr_{\ff' \draw \dist{U}_r} \left(L^{\tdist}_{\ff\ff'} \leq \epsilon\right) = 0 . \label{eq:lim}
\end{equation}
Using \cref{eq:lim} I will then show \cref{cor:ttv-invalid} by considering the two cases \(h \in \pworlds\) and \(h \not \in \pworlds\).

If we assume \(h \in \pworlds\), \cref{cor:ttv-invalid} follows  directly from \cref{cor:ttv-necessary} and \(\pworlds\) being a non-empty vector space according to \cref{thm:2} (since \(\obsvgraph\) is not \(k\)-connected).

On the other hand, if \(h \not \in \pworlds\), consider the projection of \(h\) onto \(\pworlds\) according to \(\ell\), i.e., \({f^\star = \argmin_{f}\trisk}\). Since \(\trisk\) is a Bregman divergence and \(\pworlds\) is a vector space, \cref{cor:ttv-invalid} follows then from \cref{cor:ttv-necessary}, \cref{prop:pythagorean} and the monotonicity of probability since \(L^{\tdist}_{ff^\star} \leq \trisk\). 
\end{proof}

\section{Inefficiency of scaling and benchmarks}\label{app:scaling}

\Cref{thm:2} allows to answer the \emph{scaling} question by asking how many draws from \(\sdist\) would be necessary such that all nodes are within the \(k\)-core of \(\sset\) with high probability, i.e., how many samples are needed until arriving at a valid test setting. In the following, I will discuss different ways to approximate this question.

\subsection{Scaling and the coupon collector problem}\label{app:coupon}
One way to lower bound the number of samples needed to arrive at a valid test setting would be to calculate the number of samples needed to sample each node outside the required \kcore~at least once. This is an instance of the \emph{coupon collector problem with unequal probabilities}.
In particular, let \(T_k\) be the number of draws from \(\sdist\) until we have collected \(k\) distinct nodes from \(\Set{X}_2\) for an arbitrary node in \(\Set{X}_1\). Then, it follows from \citep[Corollary 4.2]{Flajolet1992-zl} that 
\begin{equation}\label{eq:ccp}
    \textstyle 
    \E[T_k] = \sum_{q=0}^{k-1} {(-1)}^{k - 1 - q}
    \binom{m - q - 1}{m - k} 
    \sum_{|J| = q} \frac{1}{1 - P_J}
\end{equation}
where \(P_J = \sum_{j \in J} p_j\) and where \(\sum_{|J| = q}\) denotes the sum over \emph{all} subsets \(J\) of size \(q\).
However, \cref{eq:ccp} is hard to interpret and for that reason not very useful for our purposes. Moreover, \cref{eq:ccp} is not even tractable to compute at the scale that we would require for the settings considered in this paper. For instance, assume that we are dealing with a relatively small-scale domain of \(|\Set{X}| = 10^7\) entities. Since \cref{eq:ccp} requires to over all subsets of size \(k-1\), for a model of rank \(k=10\) this operation alone would require more than 
\[
    \binom{|\Set{X}|}{k-1} = \binom{10^7}{9} > 2.75 \cdot 10^{57} \text{ FLOPS} .
\]

\subsection{Proof for scaling bound in \cref{cor:inefficiency}}
Since the coupon collector problem is not computable, \cref{cor:inefficiency} considers an even weaker lower bound and asks how many samples are needed to sample an average node at least once.
For a fixed node \(x_i\), this is an instance of \emph{number of trials until first success} and follows a geometric distribution with expected value \(T_i = 1 / p_i\). Next, for a power-law distribution with \(\Pr(X > x) = u(x)x^{-\alpha}\) it holds that \(P(X = x) = u'(x)x^{-(\alpha + 1)}\) where \(u'\) is also a slowly varying function. Hence, we have 
\begin{equation}\label{eq:ccp-approx}
    T_i = \frac{1}{u'(x_i)x_i^{-(\alpha + 1)}} = \frac{x_i^{\alpha + 1}}{u'(x_i)} .
\end{equation} 
To illustrate \cref{eq:ccp-approx}, consider the following example using the Pareto distribution to instantiate \(U'\). In this case, we have
\(
T_i = x_i^{\alpha + 1} / \alpha x_{\text{min}}^\alpha
\). For a random node in \(\Set{X}\) os size \(n\), it holds then that
\begin{equation*}
    \textstyle
    \E_{i \draw \dist{U}\{1,n\}}[T_i] 
    = \frac{1}{n} \sum_{i=1}^n \frac{x_i^{\alpha + 1}}{\alpha x_\text{min}^\alpha}
    = \frac{1}{{\alpha x_\text{min}^\alpha}} \E_{i \draw \dist{U}\{1,n\}}[x_i^{\alpha + 1}]
    \geq \frac{1}{{\alpha x_\text{min}^\alpha}} \E_{i \draw \dist{U}\{1,n\}}[x_i]^{\alpha + 1}
    = \frac{1}{{\alpha x_\text{min}^\alpha}} (n/2)^{\alpha + 1}
\end{equation*}
where the inequality follows from Jensen's inequality for \(\alpha > 0\). This concludes the proof. \qedhere

\section{Experiments}\label{app:experiments}
All experiments were computed on a single \textsc{Nvidia} Volta V100 GPU and implemented using Jax~\citep{Bradbury2018-cf}, Jaxopt~\citep{Blondel2021-bx}, Numpy, and Scipy. All experiments were computed on the \textsc{MovieLens} 100k benchmark~\citep{Harper2015-po} which is available at \url{https://grouplens.org/datasets/movielens/100k/} and released under a custom license \url{https://files.grouplens.org/datasets/movielens/ml-100k-README.txt}.

\subsection{Computing possible worlds under rank constraints}\label{app:method-possworlds}
To find possible worlds that fit the observed data under a rank-constraint, I will first compute a single subspace for which we can model all observed data with highest accuracy. 
For this purpose, I am first fitting a matrix \(\mF = \mU\mV^\top\) to the observed entries under a rank constraint, i.e., via
\(\min \|\mF_{\obsvgraph} - \mY_{\obsvgraph}\|_F^2\)
where the constraint \(\rank(\mF \leq k)\) is enforced simply via \(\mU \in \R^{m \times k}\), \(\mV \in \R^{n \times k}\).
Next, let \(\mU = \mQ\mR\) be the QR decomposition of \(\mU\). Then, we know that the set of possible worlds within the subspace spanned by \(\mQ \in \R^{n \times k}\) must be of the form \(\proj_{\mQ} \cap \proj_{\obsvgraph}\) where
\begin{equation*}
    \proj_{\mQ}(\mat{M}) = \mQ \mQ^\top \mat{M}
    \quad \text{and} \quad
    {[\proj_{\obsvgraph}(\mat{M})]}_{ij} = \begin{cases}
        \ {[\mat{M}]}_{ij} & \text{if } (i,j) \in \obsvgraph \\
        \ 0 & \text{otherwise}
    \end{cases} 
\end{equation*}
are the orthogonal projections onto the column space of \(\mQ\) and the observed entries, respectively.
Furthermore, assume that we have already found \(p\) matrices that are of \(\rank\ k\) and which fit the observed entries \(\mY_{\obsvgraph}\) with high accuracy. We can then find the \(p+1\)-th matrix by minimizing the following objective:
\begin{equation}\label{eq:exfit}
    \mX = \argmin_{\mX \in \R^{m \times n}} \|\proj_{\mQ}(\mX) - \mX\|_F^2 + \|\proj_{\obsvgraph}(\mX) - \mY_{\obsvgraph}\|_F^2 - {\textstyle\sum_{i=1}^p}\|\mX - \mX_i\|_F^2 \quad \text{s.t.} \quad Y_\text{min} \leq X_{ij} \leq Y_\text{max} .
\end{equation}

Importantly, the experimental results in \cref{sec:validity,fig:experiments} hold already for only a \emph{single} subspace \(\mU\) and considering further subspaces that also explain the observed data can only increase the differences between possible worlds reported in these experimental results.

\subsection{Area over the eCDF as expected risk}\label{app:auccdf}
In the following, I will discuss how the area over the eCDF curves in \cref{fig:ecdf} correspond to the risk \(L^{\dist{U}}_{ff'}\) between these pairs of possible worlds.
In particular, let \(E\) be the random variable corresponding to the normalized absolute error of entries in possible worlds \(f\) and \(f'\). Furthermore, let \(F_E(x) = \Pr(E \leq x)\) be the CDF of \(E\). The expected error between both possible worlds (in terms of NAE) is then equivalent to the area \emph{over} the curve of eCDF, i.e., 
\begin{equation*}
    L^{\dist{U}}_{ff'} = \E_{x \draw \dist{U}}\left[\frac{|f(x) - f'(x)|}{x_\text{max} - x_\text{min}}\right] = \int_0^1 (1 - F_E(x)) dx .
\end{equation*}

\begin{figure}[h]
    \centering
    \begin{subfigure}{.5\textwidth}
        \centering
        \includegraphics[width=0.95\textwidth]{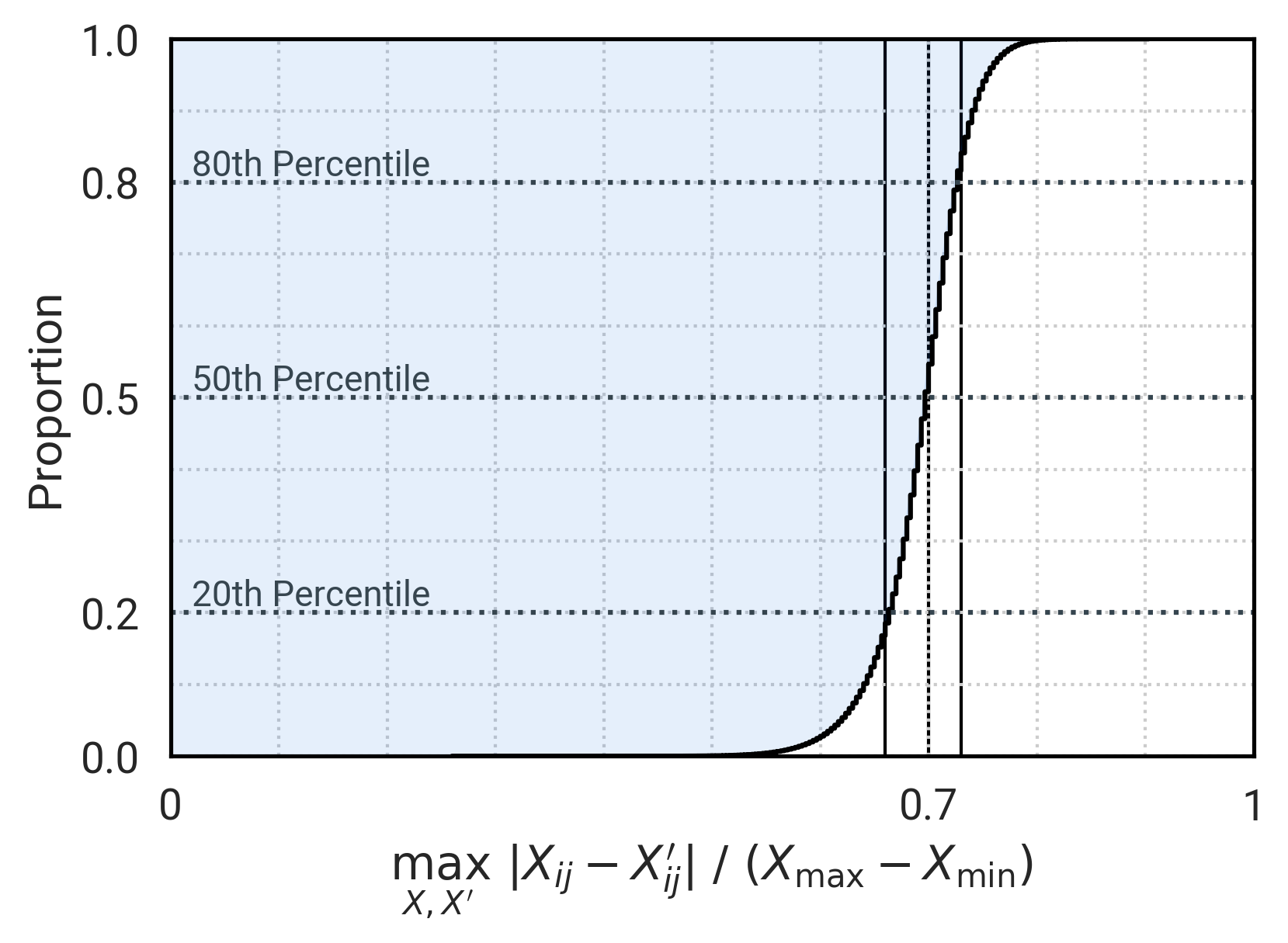}     
        \caption{eCDF of Maximum NAE }  
    \end{subfigure}%
    \begin{subfigure}{.5\textwidth}
        \centering
        \includegraphics[width=0.95\textwidth]{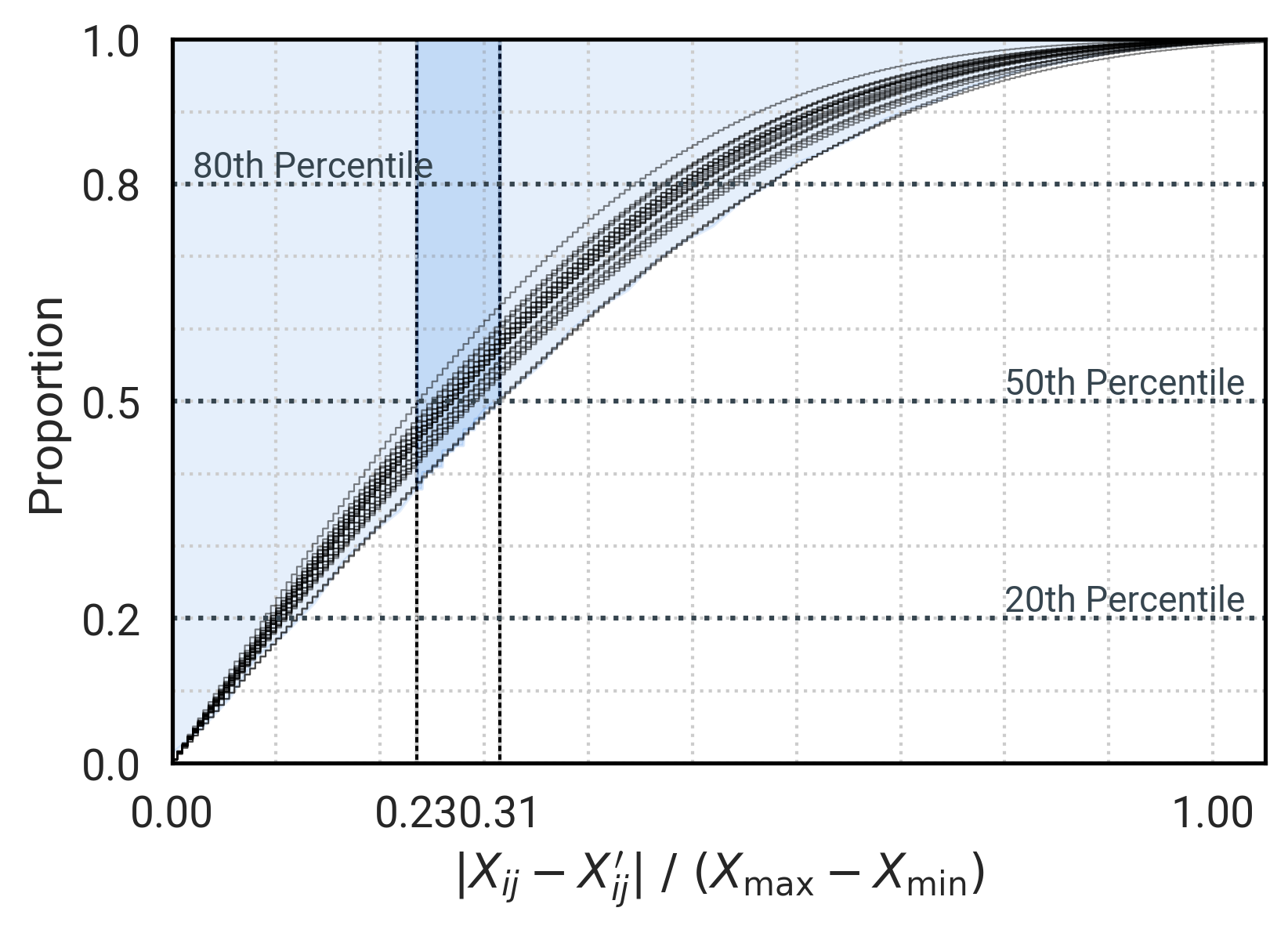}
        \caption{eCDF of Pairwise NAE}
    \end{subfigure}
    \caption{\textbf{Cora experiments}. Empirical CDF of the maximum NAE (a) and pairwise NAE (b). It can be seen that expected error (area over the curve) behaves similarly to the MovieLens dataset in the main text.}
\end{figure}

\section{Related work}\label{app:related}
In statistics,~\citet{Meng2018-km} analyzed a scaling-related question similar to this paper: Given a carefully collected survey with low response rate (small data) or a large, self-reported dataset without data curation (big data), which dataset should you trust more to estimate population averages? For this purpose, \citeauthor{Meng2018-km} introduces an Euler-formula-like identity which connects estimation quality to \emph{data quality}, \emph{data quantity}, and \emph{problem difficulty}. Similar to the results in this paper, \citeauthor{Meng2018-km} shows that data quantity is highly inefficient to overcome issues in data quality, especially sampling related issues. While related in spirit, the results in this paper go beyond the question of surveying and population averages and establish related results in the more general context of inductive inference via formalizing properties of complex social systems and their impact on validity of inferences. 

In motivation, this paper is also related to the work of~\citet{DAmour2022-xw} who study underspecification of machine learning pipelines as a cause for inference failures. In this context, a machine learning pipeline is ``the full procedure followed to train and validate a predictor''. A machine learning pipeline is then considered underspecified when it can return many distinct predictors with equivalently strong test performance. This notion of underspecification is closed related to the concepts of possible worlds and validity in this paper. 

\citet{Srebro2010-ob} studied the problem of matrix completion based on non-uniform observations such as power-laws. However, in contrast
to this work, \citet{Srebro2010-ob} assume that \(\sdist = \tdist\). The advantage of this assumption is that it leads to a much simplified learning setting in which valid inferences are indeed possible. However, as I discuss in \cref{sec:sys}, I would argue that this is not the problem that many inference settings are concerned with (and that it is questionable in a matrix completion setting as well).
Further important results in this context include \citep{Pimentel-Alarcon2016-ut} on low-rank matrix completion from deterministic samples as well as the work of \textcite{Schnabel2016-eo,Marlin2009-zy} on learning from biased samples. In contrast to these prior works, I am expanding the setting to the validity of inferences and validation, provide necessary conditions, and situate them explicitly in the context of complex social systems.

\section{Limitations}\label{app:limitations}
As most theoretical work, this work needs to make certain assumptions to make the phenomena of interest amenable to analysis. In this work, the core assumption is that samples in complex social systems follow a heavy-tailed distribution. While this is a very robust finding in social science and widely supported, as discussed in \cref{sec:sys}, it limits the results of this paper to this specific setting. For further analysis, this paper further assumes that this heavy-tailed distribution follows a regularly-varying power-law. This is again a supported assumption~\citep{Voitalov2018-jc} and allows for a clean theoretical analysis. However, as discussed in \cref{sec:sys}, it is still disputed whether samples in complex social systems actually follow this particular form. However, it is undisputed that they follow a heavy-tailed distribution, and as such, while the power-law based results might not apply exactly, their general implications are still supported.

\end{document}